\documentclass[pdflatex,sn-mathphys,iicol]{sn-jnl}% Math and Physical Sciences Reference Style
\usepackage{subfig}
\usepackage{stfloats}
\usepackage[inline]{enumitem}
\usepackage{environ}
\usepackage{refcount}

\usepackage{program}
\catcode`\|=12\relax

\usepackage{tikz}
\usetikzlibrary{calc}
\usetikzlibrary{math}

\makeatletter
\newsavebox{\measure@tikzpicture}
\NewEnviron{scaletikzpicturetowidth}[1]{%
  \def\tikz@width{#1}%
  \def\tikzscale{1}\begin{lrbox}{\measure@tikzpicture}%
  \BODY
  \end{lrbox}%
  \pgfmathparse{#1/\wd\measure@tikzpicture}%
  \edef\tikzscale{\pgfmathresult}%
  \BODY
}

\jyear{2021}%

\theoremstyle{thmstyleone}%
\theoremstyle{thmstyletwo}%

\theoremstyle{thmstylethree}%

\begin{document}
\title[KonX]{KonX: Cross-Resolution Image Quality Assessment}
%%=============================================================%%
%% Prefix	-> \pfx{Dr}
%% GivenName	-> \fnm{Joergen W.}
%% Particle	-> \spfx{van der} -> surname prefix
%% FamilyName	-> \sur{Ploeg}
%% Suffix	-> \sfx{IV}
%% NatureName	-> \tanm{Poet Laureate} -> Title after name
%% Degrees	-> \dgr{MSc, PhD}
%% \author*[1,2]{\pfx{Dr} \fnm{Joergen W.} \spfx{van der} \sur{Ploeg} \sfx{IV} \tanm{Poet Laureate} 
%%                 \dgr{MSc, PhD}}\email{iauthor@gmail.com}
%%=============================================================%%

\author*[1]{\fnm{Oliver} \sur{Wiedemann}}\email{oliver.wiedemann@uni-konstanz.de}
\equalcont{These authors contributed equally to this work.}
\author[1]{\fnm{Vlad} \sur{Hosu}}\email{vlad.hosu@uni-konstanz.de}
\equalcont{These authors contributed equally to this work.}
\author[1,2]{\fnm{Shaolin} \sur{Su}}\email{shaolin\_su@mail.nwpu.edu.cn}
\author[1]{\fnm{Dietmar} \sur{Saupe}}\email{dietmar.saupe@uni-konstanz.de}

\affil*[1]{\orgdiv{Department of Computer and Information Science}, \orgname{University of Konstanz}, \orgaddress{\country{Germany}}}

\affil[2]{\orgdiv{School of Computer Science and Engineering}, \orgname{Northwestern Polytechnical University}, \orgaddress{\country{China}}}

%%==================================%%
%% sample for unstructured abstract %%
%%==================================%%

\abstract{Scale-invariance is an open problem in many computer vision subfields. For example, object labels should remain constant across scales, yet model predictions diverge in many cases. This problem gets harder for tasks where the ground-truth labels change with the presentation scale. In image quality assessment (IQA), down-sampling attenuates impairments, e.g., blurs or compression artifacts, which can positively affect the impression evoked in subjective studies. To accurately predict perceptual image quality, cross-resolution IQA methods must therefore account for resolution-dependent errors induced by model inadequacies as well as for the perceptual label shifts in the ground truth. We present the first study of its kind that disentangles and examines the two issues separately via KonX, a novel, carefully crafted cross-resolution IQA database.
This paper contributes the following: 1. Through KonX, we provide empirical evidence of label shifts caused by changes in the presentation resolution.
2. We show that objective IQA methods have a scale bias, which reduces their predictive performance. 3. We propose a multi-scale and multi-column DNN architecture that improves performance over previous state-of-the-art IQA models for this task, including recent transformers. We thus both raise and address a novel research problem in image quality assessment.}


%%\pacs[JEL Classification]{D8, H51}

%%\pacs[MSC Classification]{35A01, 65L10, 65L12, 65L20, 65L70}

\maketitle

\begin{figure*}[t]
    \centering
    \includegraphics[width=\linewidth]{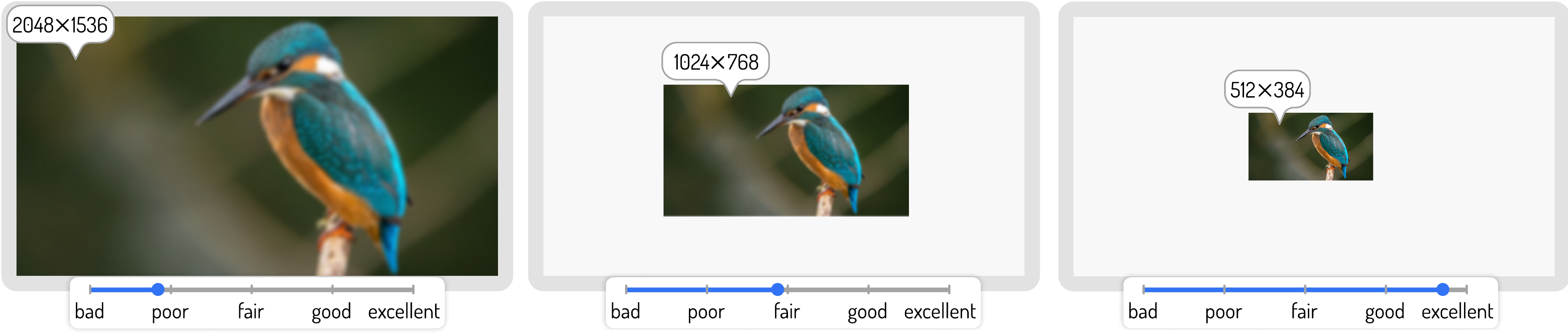}
    \caption{Scaling affects both human perception and influences IQA model predictions.}
    \label{fig:header}
\end{figure*}

\section{Introduction}\label{sec1}

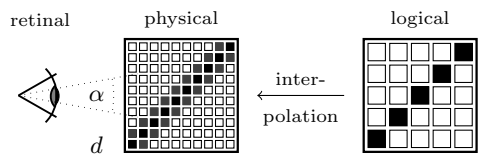
\begin{figure}[b]%{R}{0.4\textwidth}
    \centering
    \begin{scaletikzpicturetowidth}{0.8\linewidth}
    \begin{tikzpicture}[scale=\tikzscale]

    % outer bounding box
    %\draw (-1,-1) rectangle (22, 6.5);
    % draw physical pixel representation grid
    \coordinate (p) at (16, 0);
    \coordinate (lmargin) at (0.1, 0.1);
    \coordinate (rmargin) at (0.9, 0.9);
    \coordinate (upoffset) at (0,1);
    \coordinate (downoffset) at (0, -1);

    \foreach \x in {0,...,4}
        \foreach \y in {0,...,4}
            \draw ($ (p) + (lmargin) + (\x, \y) $) rectangle ($ (p) + (rmargin) + (\x, \y) $);
    \foreach \x in {0,...,4}
        \fill[black] ($ (p) + (lmargin) + (\x, \x) $) rectangle ($ (p) + (rmargin) + (\x, \x) $);
    %\foreach \x in {0,...,3}
    %    \fill[gray!40] ($ (p) + (lmargin) + (\x, \x) + (0, 1) $) rectangle ($ (p) + (rmargin) + (\x, \x) + (upoffset) $);
   % \foreach \x in {1,...,4} 
   %     \fill[gray!40] ($ (p) + (lmargin) + (\x, \x) - (0, 1) $) rectangle ($ (p) + (rmargin) + (\x, \x) + (downoffset) $);
    \draw[line width = 0.3mm] ($ (p) - (0.1, 0.1)$) rectangle ($ (p) + (5.1, 5.1) $); 
    
    \node[align=center] at (7.5, 6) {\footnotesize physical};

    % virtual pixel representation grid
    \coordinate (v) at (5, 0);
    \tikzmath{\s  = 0.5;} % scale factor
    \coordinate (lmargin) at (0.075, 0.075);
    \coordinate (rmargin) at (0.425, 0.425);
    \coordinate (upoffset) at (0, 0.5);
    \coordinate (downoffset) at (0, -0.5);

    \foreach \x in {0,...,9}
        \foreach \y in {0,...,9}
            \draw ($ (v) + (lmargin) + (\s*\x, \s*\y) $) rectangle ($ (v) + (rmargin) + (\s*\x, \s*\y)$);
    \foreach \x in {0,...,9}
        \fill[black] ($ (v) + (lmargin) + (\s*\x, \s*\x) $) rectangle ($ (v) + (rmargin) + (\s*\x, \s*\x) $);
    \foreach \x in {0,...,8}
        \fill[darkgray] ($ (v) + (lmargin) + (\s*\x, \s*\x) + (upoffset) $) rectangle ($ (v) + (rmargin) + (\s*\x, \s*\x) + (upoffset) $);
    \foreach \x in {1,...,9} 
        \fill[darkgray] ($ (v) + (lmargin) + (\s*\x, \s*\x) + (downoffset) $) rectangle ($ (v) + (rmargin) + (\s*\x, \s*\x) + (downoffset) $);
    \draw[line width = 0.3mm] ($ (v) - (0.1, 0.1)$)rectangle ($ (v) + (5.1, 5.1) $); 
    
    \node at (18.5, 6) {\footnotesize logical};

    \coordinate (eye) at (0, 2.5);
    % upper and lower straight lines
    \draw[line width=0.25mm] (eye) -- ($ (eye) + (1.75, 1) $);
    \draw[line width=0.25mm] (eye) -- ($ (eye) + (1.75, -1) $);
    % front arc
    \pgfmathsetmacro{\eyeRadius}{1.75}
    \pgfmathsetmacro{\outer}{45} % outer angle
    \pgfmathsetmacro{\irisSize}{45} % iris size (as an angle)
    \pgfmathsetmacro{\irisRadius}{0.5} %iris radius

    \draw[fill=gray, line width=0.25mm] (eye)++(15:\eyeRadius) arc (-\irisSize+180:\irisSize+180:0.4*\eyeRadius);
    \draw[fill=black] (eye)++(-15:\eyeRadius) arc (-\irisSize:\irisSize:0.35*\eyeRadius);
    \draw[line width=0.25mm] (eye)++(\outer:\eyeRadius)++(0.01,0) arc (\outer:-\outer:\eyeRadius);

    % dashed line from iris to pixel
    %\draw[line width=0.15mm, dashed] (eye)++(0:\eyeRadius) -- (5.5, 2.5);
    %\draw[line width=0.15mm] (5.4, 2.4) -- (5.6, 2.6);
    %\draw[line width=0.15mm] (5.4, 2.6) -- (5.6, 2.4);
  
    % viewing distance d
    \draw[line width=0.15mm] (eye)++(0:\eyeRadius)++(0, -3) -- ($(5.5, 2.5)+(0, -3)$);
    \draw[line width=0.15mm] (eye)++(0:\eyeRadius)++(0, -3) -- ($(eye)+(0:\eyeRadius)+(0, -3)+(0,0.1)$);
    \draw[line width=0.15mm] ($(5.5, 2.5)+(0, -3)$) -- ($(5.5, 2.5)+(0, -3)+(0,0.1)$);
    \node at (3.6, 0.25) {\small $d$}; 
    
    % visual angle alpha 
    \draw[line width=0.1mm, dotted] (eye) -- (5.5, 3.5);
    \draw[line width=0.1mm, dotted] (eye) -- (5.5, 1.5);
    \draw[line width=0.1mm, dotted] (eye)++(10:2.5*\eyeRadius)++(0.01,0) arc (10:-10:2.5*\eyeRadius);
    \node at (3.6, 2.5) {\small $\alpha$};
    
    % dotted lines going from the eye to the left
    %\draw[line width=0.1mm, dotted] (eye) -- (-2.75, 3);
    %\draw[line width=0.1mm, dotted] (eye) -- (-5, 1.5);
    
    % perceived image on the left
    \coordinate (p) at (-7, 0);
    \tikzmath{\s  = 0.5;} % scale factor
    \coordinate (lmargin) at (0.075, 0.075);
    \coordinate (rmargin) at (0.425, 0.425);
    \coordinate (upoffset) at (0, 0.5);
    \coordinate (downoffset) at (0, -0.5);
    %\foreach \x in {0,...,9}
    %    \foreach \y in {0,...,9}
    %        \draw ($ (p) + (lmargin) + (\s*\x, \s*\y) $) rectangle ($ (p) + (rmargin) + (\s*\x, \s*\y)$);
   % \foreach \x in {0,...,9}
        %\fill[black] ($ (p) + (lmargin) + (\s*\x, \s*\x) $) rectangle ($ (p) + (rmargin) + (\s*\x, \s*\x) $);
    %\foreach \x in {0,...,8}
   %     \fill[black] ($ (p) + (lmargin) + (\s*\x, \s*\x) + (upoffset) $) rectangle ($ (p) + (rmargin) + (\s*\x, \s*\x) + (upoffset) $);
    %\foreach \x in {1,...,9} 
    %    \fill[black] ($ (p) + (lmargin) + (\s*\x, \s*\x) + (downoffset) $) rectangle ($ (p) + (rmargin) + (\s*\x, \s*\x) + (downoffset) $);
    %\draw[line width = 0.3mm, dotted, rounded corners] ($ (p) - (0.1, 0.1)$)rectangle ($ (p) + (5.1, 5.1) $); 

    \node at (1, 6) {\footnotesize retinal};

    \draw[->] (15, 2.5) -- (11, 2.5);
    \node at (13, 3.25) {\footnotesize inter-};
    \node at (13, 1.5) {\footnotesize polation};

\end{tikzpicture}
    \end{scaletikzpicturetowidth}
    \vspace{15pt}
    \caption{The term \textbf{resolution} can be ambiguous. In this paper \textbf{we use it for the \textit{logical} image size of $w\times h$ pixels}. Presenting an image on a screen, possibly interpolated, yields a \textit{physical} resolution, which defines the image's spatial dimensions and pixel density. What matters most for the human visual system is the perceivable \textit{angular} resolution, which depends on the physical pixel density on the screen, the distance $d$ to the screen, and the minimal discernible angle $\alpha$. The result is a representation of the image on the retina, which in turn evokes an impression in the visual cortex. }% (of visual quality) in the brain.}
    \label{fig:resolutions}
\end{figure}

The discipline of image quality assessment (IQA) aims to model how humans perceive the quality of digital images. Recent no-reference (NR-)IQA algorithms predict quality scores for a given input without a pristine reference. They perform well when tested on the same domain as they were trained on; however, model performance drops when cross-tested on different datasets \cite{koniq10k, su2021koniq++, talebi2018nima}. 
We hypothesize that this decrease in performance is caused by two factors: a lack of \textit{cross-resolution generalization} by the models and \textit{domain shifts} across datasets. The latter is concerned with image contents and differences in the distributions of distortion types, combinations, and their severity.
We aim to isolate the first factor, which is also known as the \textit{cross-resolution problem}, for image quality assessment. To this end, we created a first-of-its-kind dataset that provides a reliable benchmark for cross-resolution IQA. By \textit{resolution} we mean \textit{image size in pixels}, which is to be distinguished from \textit{resolution as pixel densities}. On a display these are expressed in terms of dots or pixels per inch (DPI/PPI), whereas on the viewer's retina a notion of angular resolution is better suited, as illustrated in Fig. \ref{fig:resolutions}.

Previous works in NR-IQA \cite{bosse2016deep, koniq10k, li_norm-in-norm_2020, su2021koniq++, talebi2018nima} assumed that the quality ratings of images gathered at one presentation resolution are valid at other resolutions as well. This is not the case. We subsequently show that perceived quality varies with the presentation resolution. When comparing images across resolutions, we get only a 0.93 Spearman rank-order correlation coefficient (SRCC) between their mean opinion scores (MOS) when the scale ratio is 4:1, compared to a 0.97 SRCC when it is 2:1.
Reliable IQA for modern high-resolution images is desirable, as it could pave the way for its wider application beyond academic research. Existing NR-IQA methods do not perform well in cross-resolution settings. This is in part because existing IQA databases are annotated at comparatively low resolutions and because the prevalent approach is to train and test them on images that were resized to the same scale~\cite{koniq10k, li_norm-in-norm_2020, su2021koniq++}. 

Some existing IQA datasets (e.g. \cite{ying2020patches}) contain images of various resolutions. However, there is none that was \textit{annotated} at multiple resolutions, but the images were either scaled to a fixed presentation size or presented in their native resolution with different spatial sizes on screen. Rigorous cross-resolution comparisons on the same content were thus not possible.
To address these limitations, we created \textit{KonX}, a database in which the same image contents were annotated at multiple presentation scales. It serves as the first cross-resolution benchmark and allows to test quality predictors at multiple resolutions.%, which was not possible in a rigorous fashion on previous datasets.% Previously, it was impossible to properly validate architecture performances across scales. %We contribute at every stage of the process, from the database via models enhancements to the validation strategy.

\subsection{Contributions of this Work}
We introduce a novel problem, create a database that allows us to approach it for the first time, propose a DNN architecture that surpasses the state of the art, and add validation considerations that allow proper comparisons of cross-resolution model performances. In greater detail:

\subsubsection{A Novel Problem} The cross-resolution problem in NR-IQA arises by distinguishing between \textit{cross-content} and purely \textit{cross-resolution} predictions. The latter approach removes the confounding variable of image content from our experiments. This has not been studied before: previous IQA datasets only provided one annotation resolution per content and particularly for crowdsourced studies it is often unclear how well the annotation resolution was controlled for in the actual studies \cite{ghadiyaram2015live, ponomarenko2015image, koniq10k, ying2020patches}. 

%and some did not properly control for the annotation resolution at all, particularly in crowdsourced studies.

\subsubsection{A New Dataset} \textit{KonX} shows that the label shift is significant and that current NR-IQA models are unable to account for it.
We took multiple measures to achieve precise annotations:
\begin{enumerate}[label=\roman*)]
\item By inviting expert freelancers as participants.
\item By conducting a longitudinal study in which all items were rated twice, which provides valuable information about participant reliability, self-consistency and attention levels.
\item By controlling the presentation size. Our interface renders logical pixels 1:1 to screen pixels, which was not ensured for any previous NR-IQA dataset.
\end{enumerate}

\subsubsection{A DNN Architecture Proposal} In multi-column architectures, weights are usually shared between columns to limit the capacity and prevent overfitting. We employ a transfer-learning backbone in a multi-column architecture \textit{with individual weights} that still does not overfit. The key is to feed different resolutions to each column and create a bottleneck before combining per-column features. We also integrate information from multiple levels of the network, i.e., from all psedo-repeated modules of the \texttt{EfficientNet} backbone. These scale-variant features further improve the cross-resolution performance. 

\subsubsection{Validation Considerations}
Absolute score prediction is crucial in cross-resolution IQA, as the ground-truth MOS changes with the image resolution. By validating NR-IQA methods on absolute errors \textit{and} rank correlation to ground-truth, we demonstrate the limitations of singular metric choices. Our model outperforms recent competition in cross-database and cross-resolution comparisons w.r.t. both metrics.

\begin{figure*}[t]
\subfloat[$2048\times1536$]{\includegraphics[width=0.225\linewidth]{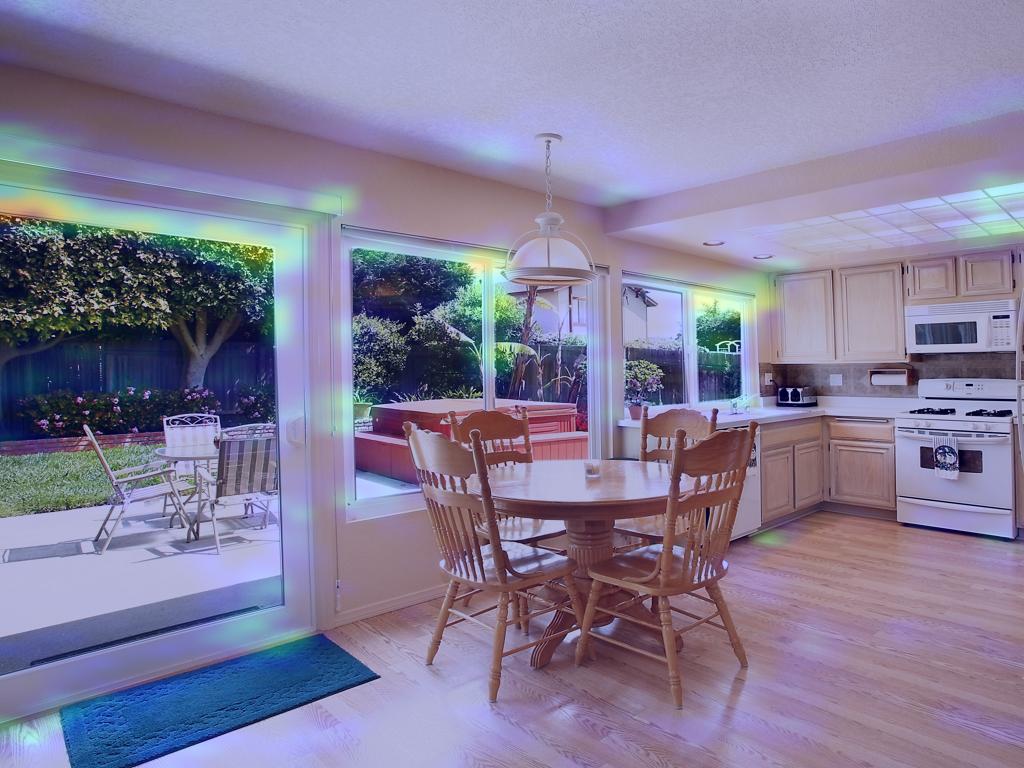}} \hfill
\subfloat[$1024\times 768$]{\includegraphics[width=0.225\linewidth]{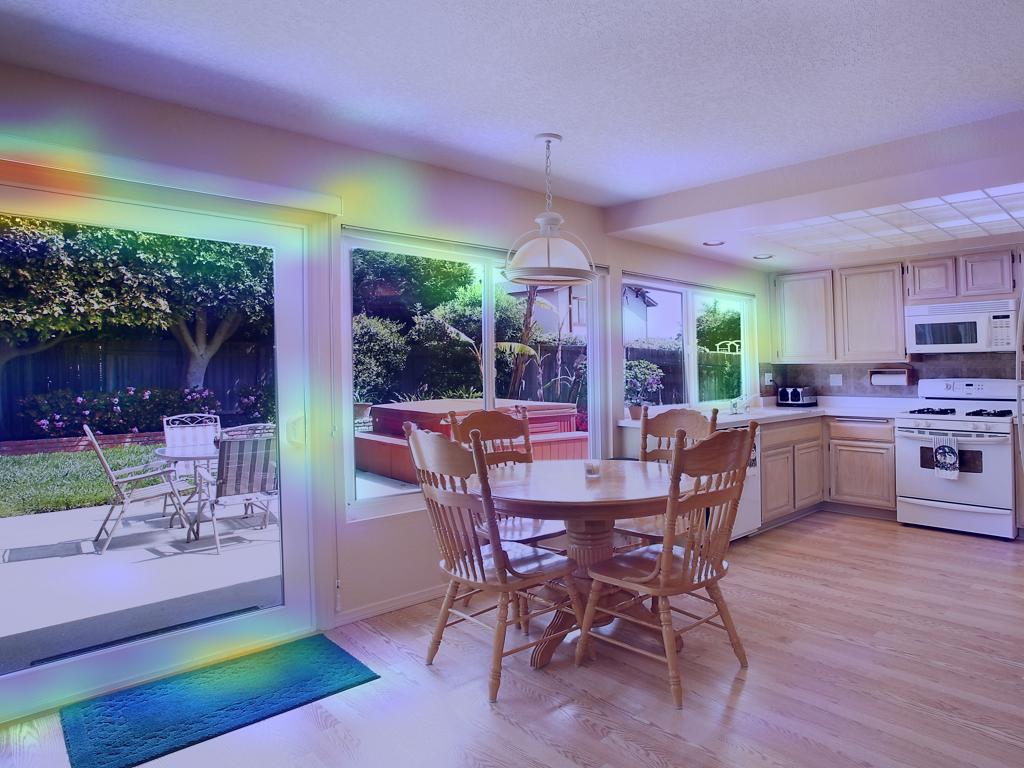}} \hfill
\subfloat[$512\times 384$]{\includegraphics[width=0.225\linewidth]{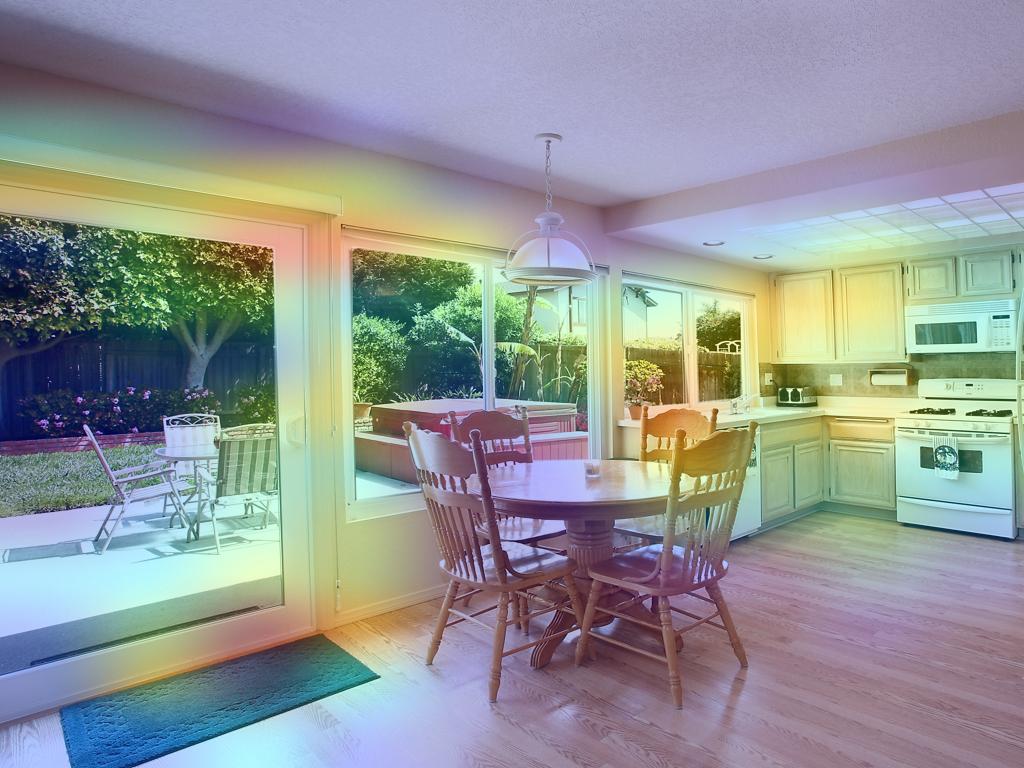}} \hfill
\subfloat[$256\times192$]{\includegraphics[width=0.225\linewidth]{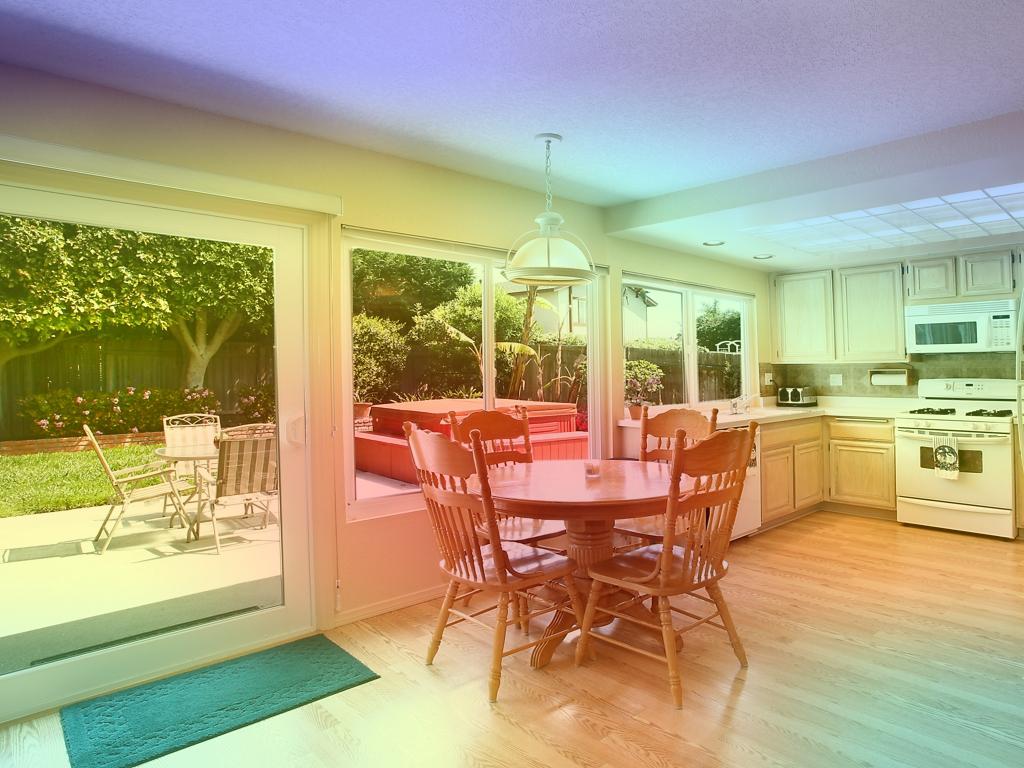}} \\
\centering
\caption{The cross-resolution problem: Grad-CAM \cite{selvaraju2017grad} heatmaps depict aberrant regions-of-interest for the top predicted class of an InceptionResNetV2 \cite{szegedy2017inception}. Analogous difficulties in CNN-based IQA methods are even more delicate, as perceptual quality varies with scale, unlike object class labels.}
%Cross-resolution discrepancy in object recognition: When exposing an InceptionResNetV2 \cite{szegedy2017inception} classifier to \texttt{445846327.jpg} from KonIQ-10k \cite{koniq10k} at different resolutions, one observes diverging Grad-CAM \cite{selvaraju2017grad} ROIs for the top-1 predictions. Analogous difficulties in CNN-based IQA methods are even more delicate, as quality is not scale-invariant, unlike object class labels.}
\label{fig:gradcam}
\end{figure*}

\section{Related Work}
\label{sec:related}

\subsection{IQA Models}

Perceptual quality prediction evolved from statistical methods \cite{wang2004image,sheikh2005visual} to an application area of deep learning. Most approaches crop or scale their input to a fixed, usually small resolution
\cite{kang2014convolutional,bosse2016deep,wiedemann2018disregarding,talebi2018nima,zhu2020metaiqa,yang2019sgdnet,pan2018blind,golestaneh2022no}.
We aim to make IQA applicable at resolutions that are relevant in practice and focus on \textit{no-reference} or \textit{blind} IQA models, which take only the distorted image as an input and predict a quality score directly \cite{hosu2020koniq, su2021koniq++,li_norm-in-norm_2020}. In comparison to \textit{full-reference} IQA scenarios, where one has access to both the distorted image and a usually pristine original, the performance of NR-IQA methods in cross-resolution and cross-database tests is significantly reduced, especially on \cite{ghadiyaram2015live,ying2020patches}. This is due to a more general problem in computer vision: scale variance \cite{van2017learning}, which in this case manifests itself as the \textit{cross-resolution problem}.  

Regarding model architectures we took inspiration from successful and recent works, of which some already leaned towards improving robustness against input scale variance.
Aggregating activations of multiple layers of pre-trained CNNs through a second network for example has shown success in image aesthetics assessment (IAA) \cite{hii2017multigap,hosu2019effective}. This inspired us to employ multi-level spatially pooled (MLSP) features in our proposed architecture as well. We noticed that CNNs \cite{koniq10k} still perform well on KonIQ even in comparison to transformer-based architectures \cite{ke2021musiq,golestaneh2022no}, in this case with SRCCs of 0.921 (KonCept-512) vs. 0.916 (MUSIQ)  and 0.915 (Golestaneh et al.). One hypothesis is that the use of both multi-scale inputs and multi-level features would be beneficial for cross-resolution prediction. 
%We noted that popular \cite{ke2021musiq} and recent \cite{golestaneh2022no} transformer-based approaches are still outperformed on KonIQ-10k \cite{koniq10k} by the model published alongside the database, which is ignored in their comparisons.
Furthermore, it is unclear if transformers perform better in IQA than traditional CNNs, especially so for cross-resolution tasks.

Some works on full-reference IQA \cite{wang2003multiscale,Temel2015multi} integrate information from downscaled versions of their input internally. However, they're only evaluated on predictions for a single fixed resolution, so they don't approach the problem of resolution-dependent scores.
NR-IQA models additionally have to intrinsically encode both the knowledge about visual distortions and their connection to the image resolution. Only a few attempts on multi-scale approaches in NR-IQA \cite{ke2021musiq,you2021transformer} have been made. %, the latter source seems to report cross-resolution performance on the training dataset. % As there are no previous datasets annotated at multiple resolutions, \textit{KonX} enables us to thoroughly investigate this problem in IQA for the first time.
We considered adding explicit information about the scale similar to \cite{ke2021musiq}, but \cite{graziani2021ScaleInvarianceState} has shown that CNNs can  infer the input dimensions by using the 0-padding that is added to images before convolution kernels are applied.
Another factor to consider is the prediction target. Three main types are found in the IQA literature: a single rating per image \cite{koniq10k}, the distribution of ratings from multiple annotators \cite{talebi2018nima,ke2021musiq} and scale-free rankings rather than absolute ratings \cite{li_norm-in-norm_2020,liu2018EndtoEndBlindQuality}.
This work aims to predict a single rating per image, as accurately as possible across resolutions.

The MSE loss is a reasonable choice due to its characteristics when training for absolute scores. In our experiments, it did not perform worse than alternatives even when the evaluation metric is Spearman's rank correlation coefficient between predictions and ground-truth ratings \cite{koniq10k}, as commonly used in IQA. This applies to all three types of losses previously mentioned, including the scale-free rating loss introduced by Li et al. \cite{li_norm-in-norm_2020}. The latter work's improved performance seems to be primarily due to the choice of training resolution, rather than the loss itself, and though it appears to converge faster in the early epochs, there is no clear overall advantage compared to the MSE.

\subsection{Scale Generalization}% and Architecture Considerations}

We incorporated works on scale generalization and transfer-learned CNNs in order to build a model that accurately predicts quality scores across resolutions. The base architecture, usually a pre-trained (e.g., on ImageNet) feature extractor, is a key choice. We expect newer architectures to generally improve performance, but multiple factors play a role.
ImageNet CNNs are usually trained at small resolutions, many at $224\times224$ pixels, up to %$600\times600$ EfficientNet-B7 \cite{tan2019efficientnet} or 
$800\times800$ for EfficientNet-L2 \cite{touvron2020fixing}. Pre-training on such small resolutions might limit the performance in large-resolution IQA.
InceptionResNet-v2 was applied successfully in IAA \cite{hosu2019effective} on AVA \cite{murray_ava_2012}, an aesthetics database that contains images of various resolutions (up to $800\times800$). It outperformed other proposals in the past years since its introduction, which raises the question: \emph{what makes this particular architecture more suitable for cross-resolution tasks?}

Recent quality and aesthetics models \cite{li_norm-in-norm_2020,su2020BlindlyAssessImage,hosu2019effective} combine activations from multiple layers of pre-trained backbone models.
Later-stage layers of ImageNet models usually represent abstract, scale-invariant concepts \cite{graziani2021ScaleInvarianceState}, whereas
earlier layers tend towards scale-dependent features.
IQA depends on both, e.g., object classes and pixel-level distortion patterns. This explains the benefit of integrating information from multiple layers of an object classification network for IQA.
%We propose a model that considers multi-level features, extending \cite{lin2020deepfl,hosu2019effective} to a new architecture.

CNNs trained on a single resolution \cite{graziani2021ScaleInvarianceState,touvron2020fixing} exhibit scale-wise overfitting, which can be mitigated by multi-resolution ensembles \cite{vannoord2017LearningScalevariantScaleinvariant}.
Multi-column architectures have shown success in crowd-counting \cite{Kang2018crowd,zhang2016SingleImageCrowdCounting,walach2016LearningCountCNN,onororubio2016Perspective}, which involves varying object scales within single images. Again, this integrates information from multiple scales: \cite{Kang2018crowd} feed rescaled images to a shared-weight CNN column.
Most crowd-counting works use directly trained custom architectures for the task, but we consider pre-trained networks as columns in hopes that they can jointly handle different scales.

\subsection{Databases}

IQA datasets are classified into two types: those with \emph{artificially} distorted images and those with \emph{authentically} distorted images. The former are derived from pristine originals by applying distortions of various types and magnitudes, either single or in combinations \cite{sheikh2005live,ponomarenko2015image,liu2014cid,lin2019kadid}.
This class has been criticized for lacking diversity due to the comparatively small sets of source images and the limited variety of distortions.
Models trained on it have poor generalization to new impairments \cite{lin2020deepfl}. % and are less practical.
On the other hand, authentically distorted IQA databases are usually sampled directly from online photography communities. The images are affected by mixtures of naturally occurring distortions.
The state of the art for general authentically distorted IQA databases is currently \emph{KonIQ-10k} \cite{hosu2020koniq}, with 10,073 images. 
\textit{SPAQ} \cite{Yumin2020Perceptual-SPAQ} is the largest domain-specific \textit{authentic} dataset with 11,125 images taken with smartphone cameras.

Another subclass of databases focuses on local image quality, a concept introduced by \textit{KonPatch-30k} \cite{wiedemann2018disregarding} and extended through \textit{Paq-2-Piq} \cite{ying2020patches}.
They allow to compare the quality of patches with that of the entire image, which generalizes the concept of a global MOS to local image quality.

However, using only these existing IQA datasets, one \emph{cannot} reliably study the cross-resolution problem. Though there are datasets that annotate different images, or crops thereof, at different resolutions, such as \textit{SPAQ} \cite{Yumin2020Perceptual-SPAQ} and \textit{Paq-2-Piq} \cite{ying2020patches}, no dataset so far annotated the same image contents at multiple presentation resolutions. This means neither the subjective perceptual shifts across resolutions, nor the reason why IQA models perform poorly in cross-resolution (and cross-dataset) tests is studied thoroughly. %OLD PHRASING: one cannot be certain whether models perform poorly in cross-resolution tests due to a change in resolution or image content.

Our proposed dataset, \textit{KonX}, allows to properly validate the cross-resolution performance of IQA models for the first time by comparing predictions versus three resolution-specific mean opinion scores.
We conducted a crowdsourcing-based user study to obtain subjective ratings specifically for the cross-resolution testing. We anticipate that our work will pave the way for new directions in image quality research.

% Image quality research is driven by subjective ratings obtained in user studies. Supervised machine learning requires huge amounts of data, leaving crowdsourcing the only viable labeling option. Due to a lack of high-resolution screens in the crowd, annotating large full-sized images is not possible. This also emphasizes the utility of IQA models that perform well across resolutions in order to make automated assessments on high-resolution images feasible.% Bridging the resolution gap between current camera sensors and IQA datasets is currently aggravated by mere limitations in large-scale data acquisition. 

\subsection{Subjective Factors in QoE}
\label{sec:subjective-factors-in-qoe}
    
Previous studies in which existing IQA databases were annotated did not consider well-known aspects of \textit{quality of experience} (QoE). 
Reiter et al. \cite{reiter2014FactorsInfluencingQuality} introduced three classes of influence factors (IFs) in this regard:
\emph{Human} IFs affect the lower-level (visual acuity, age, mood, etc.) and higher-level (cognitive processes, personality traits, expectations, etc.) perception of quality. 

\emph{System} IFs are related to content, network, and device aspects (screen resolution, display size, etc.), while \emph{context} IFs are affected by the environment (temporal, social, technical peculiarities, etc.). Many Reiter IFs are difficult to study, especially in crowdsourcing, where control mechanisms are lacking and self-reports can be unreliable.
Several studies \cite{moorthy2012VideoQualityAssessment,gong2014ImpactsAppearanceParameters,rehman2015DisplayDeviceadaptedVideo,zou2016PerceivedImageQuality,sotelo2017SubjectiveVideoQuality,kara2019ComparisonHDUHD, saad2015impact} report on the influence of the display device (System IF) on the perceived quality, especially regarding device characteristics.

The \textit{visual resolution} \cite{rossiLimitsVisualResolution} of an image presentation imposes a limit on the pixels that are discernible by the human visual system. 
It depends on the display size, its physical resolution, the mapping from virtual- to physical pixels, the viewing distance, and finally, the viewer's physiological capabilities, as shown in Fig. \ref{fig:resolutions}.
Opposing effects can occur when altering the visual resolution:

\begin{itemize}
    \item Presenting a pristine image at a higher visual resolution can increase its perceptual quality, as additional details become visible \cite{kim2008FactorsAffectingPsychophysical}.
    \item A reduced visual resolution of a degraded image can mask impairments, which in turn can \textit{also} increase perceptual quality.
\end{itemize}

\noindent
Both effects play a role in quality assessment but have not been considered in previous works, let alone handled consistently.
Moorthy et al. \cite{moorthy2012VideoQualityAssessment} presented videos centered on mobile screens, while Gong et al. \cite{gong2014ImpactsAppearanceParameters} resized images to ensure a constant physical size. On the other hand, Zou et al. \cite{zou2016PerceivedImageQuality} and Kara et al. \cite{kara2019ComparisonHDUHD} opted for full-screen, rescaled as needed.
The source images were not always the same size as the screen resolution.

Rehman et al. \cite{rehman2015DisplayDeviceadaptedVideo} did not state what the presentation size was, but it can be assumed to be full-screen. None of the authors mention possible discrepancies between the virtual and physical resolutions. This is relevant nowadays, especially when presenting images in browser-based user interfaces due to the reliance on rendering at virtual resolutions that are smaller than the physical ones. Apple Retina displays, for example, have ratios between the physical and virtual resolution up to 3:1. We consider these aspects in our study and control for them as much as possible.

The viewing distance (Human/Context IF) between participants and the screen was considered before. Studies involving 4K TVs \cite{kara2019ComparisonHDUHD} deemed it essential to be controlled, less so those on mobile and desktop devices \cite{moorthy2012VideoQualityAssessment,zou2016PerceivedImageQuality}. The latter emphasizes the freedom to choose one's preferred viewing distance to best express natural behavior instead of enforcing strict, possibly awkward or even uncomfortable scenarios, e.g., chin rests.
Following this line of reasoning, we did not expect participants in our study to maintain a fixed viewing distance. It is not only difficult to enforce this in crowdsourcing, but feeling uncomfortable might reduce the participants' ability to focus on the assessment task and negatively affect their judgments.

\section{The KonX Database}
\label{sec:dbcreation}

Our novel cross-resolution IQA database \textit{KonX} was annotated with subjective quality scores at three presentation resolutions. It is primarily intended as a benchmark for IQA models. With its emphasis on annotation reliability, it allows for the first time to investigate the relationship between perceived quality and scale.

\begin{figure*}[t]
    \centering
    \includegraphics[width=\linewidth]{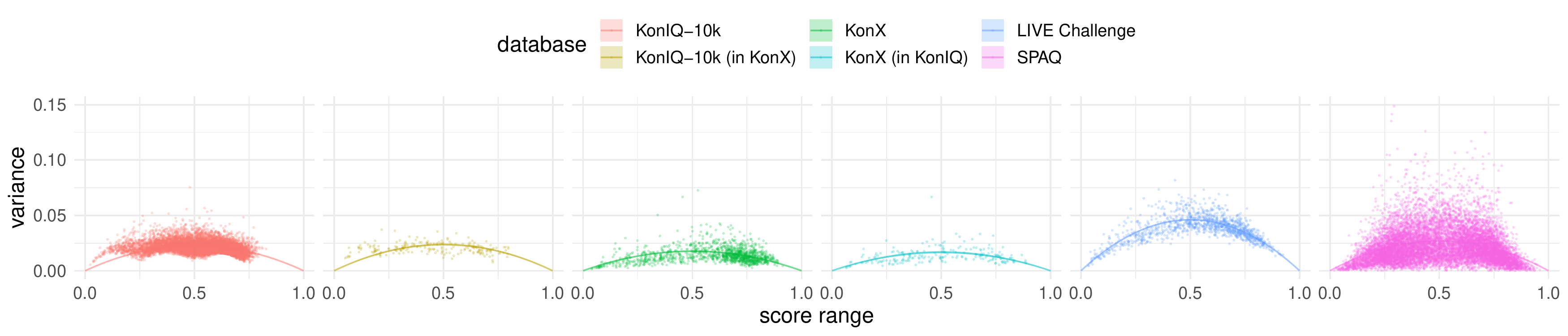}
    \caption{Variance versus MOS of authentically distorted, crowdsourced datasets. The SOS-hypothesis $a$ values for KonX, KonX scores at $1024\times768$ for the subset of images sampled from KonIQ-10k, KonIQ-10k, KonIQ-10k scores for the subset of images sampled for KonX, Live Challenge, and SPAQ are 0.071, 0.067, 0.091, 0.095, 0.184, and 0.107 respectively. The 95\% confidence interval for $a$ is indicated by the shaded region around the main curve.
    \label{fig:sos-plot-authentic}}
\end{figure*}

\subsection{Introduction Overview}

KonX consists of 210 images from \textit{Flickr}\footnote{\label{flickr}\url{https://flickr.com}}, which were already included in \textit{KonIQ-10k} \cite{koniq10k}, and another 210 images from \textit{Pixabay}\footnote{\label{pixabay}\url{https://pixabay.com}} to supplement the high-quality range.
The images were sampled using a stratified approach based on discretized metadata and other image properties. We aimed to diversify both their perceptual quality levels and contents.
We center-cropped all Pixabay candidates, and smart-cropped~\cite{hosu2020koniq} the KonIQ-10k original images to an aspect ratio of 4:3. These were then downsampled using the Lanczos-interpolation to three resolutions: $2048 \times 1536$px, $1024 \times 768$px and $512 \times 384$px.

Eighteen freelancers\footnote{\url{http://freelancer.com}} with a professional background in photography or graphics design rated each image twice at each resolution.
The study participants were thoroughly screened for their ability to detect image defects. We deployed a custom web interface that ensures a 1:1 rendering of virtual image pixels to physical screen pixels without scaling, thus displaying the lower-resolution images at a smaller spatial size. 
This experimental setup resulted in $45360$ annotations of $420$ image sources at three resolutions. We now explain and justify the choices behind \textit{KonX} in detail. The most important facts are summarized in Table \ref{tab:KonX_facts}.
%\subsection{Annotation Reliability Analysis}
%An indicator of the reliability is presented in Figure \ref{fig:ICCs-databases-more}. %We conducted a thorough analysis and comparison of the annotations provided in KonX, which is included as a technical report in the supplementary materials.
%\textcolor{red}{\textit{This is only a brief summary of the KonX database. Please refer to the supplementary material for a detailed report on our subjective study and thorough reliability analysis.} }

\begin{table}[h]
    \centering
    \caption{KonX: A Cross-Res. IQA Benchmark}
    \resizebox{\linewidth}{!}{
    \begin{tabular}{l|l}
        \hline
        Sources & \textit{Flickr} (KonIQ-10k) and \textit{Pixabay}\\ \hline
        \#Images       & 210 from each source\\ \hline
        Resolutions &  $2048\times1535$px, $1024\times 768$px, $512\times384$px\\ \hline
        Participants & 19 in the full study\\ \hline
        Annotations &  2 per image at each resolution, 45360 in total\\
        \hline
    \end{tabular}
    }
    \label{tab:KonX_facts}
\end{table}

\subsection{Content Preparation}
When creating an image database, one of the main goals is to reduce potentially unknown biases, which stem from shared characteristics among images.
This can be mitigated by enforcing \textit{diversity} through adequate sampling strategies. Similar goals have been set for previous IQA \cite{koniq10k} and VQA \cite{hosu2017konstanz} datasets.
We incorporated several means to diversify \textit{KonX} with respect to perceptual quality as the primary attribute as well as auxiliary aspects such as image content. %The approach was as follows:

\subsubsection{Data Sources}
We sampled from two online photography platforms: \textit{Flickr}\footnotemark[\getrefnumber{flickr}] and \textit{Pixabay}\footnotemark[\getrefnumber{pixabay}]. All candidate images from \textit{Flickr} were already included in \textit{KonIQ-10k} \cite{koniq10k}, which provides preexisting MOSes for comparison. This set was augmented with content from \textit{Pixabay}, which offers mostly high-resolution images. The goal was to supplement the high-quality range in which \textit{KonIQ-10k} is lacking. 

\subsubsection{Resolution and Aspect Ratio}
Candidate images from both sources had be larger than $2048\times 1536$px and have aspect ratios between $[1.315, 1.785]$ to retain similarity. We extracted image content at $2048\times1536$px, $1024\times768$px and $512\times384$px by cropping the original images to an aspect ratio of 4:3. We cropped the center part of the image for Pixabay, and used the smart-cropping \cite{hosu2020koniq} procedure for KonIQ-10k. The crops were then downsampled to $2048\times1536$px and the aforementioned lower resolutions using Lanczos interpolation. On the \textit{Flickr} subset, this enforced identical image portions as published in the \textit{KonIQ-10k} dataset at $1024\times768$px.

\subsubsection{Stratified Attribute Sampling}
Our sampling strategy relies on stratified discrete attributes, for which \textit{Flickr} and \textit{Pixabay} provide different tags and metadata.
The occurrence frequencies of unique values were treated as ``levels'', over which we aimed for uniformity.
We additionally included machine tags from \cite{yfcc100m} for the \textit{Flickr} candidates.
The pre-existing MOSes from KonIQ-10k were quantized into equal-length bins to fit into our discrete approach. For the \textit{Pixabay} candidates, we considered the camera model, user-assigned tags and incorporated \textit{normalized favorites} $\widetilde F(I)$. This measure is calculated as follows, where $F(I)$ is the number of ``favorites'' that image $I$ received on the \textit{Pixabay} platform and $V(I)$ is the total number of times it was \textit{viewed}:
\begin{equation}
\widetilde F(I) = \ln(F(I) + e)/\ln(V(I) + e) \label{normfavs}
\end{equation}
On the admissible $7818$ \textit{Flickr} and $757.016$ \textit{Pixabay} images, we iterated the following procedure, thereby sampling 210 images from each source:
\begin{enumerate}[label=\roman*)]
    \item Randomly select an attribute.
    \item Randomly select one of its available ``levels".
    \item Keep the images corresponding to this choice.
    \item On this subset, continue alike with step i)
\end{enumerate}
After all attributes have been considered, the procedure either returns a single image or a set of images. In the latter case, we chose one image at random.

\subsection{Subjective Annotation Study}
In order to establish a benchmark that allows meaningful comparisons across resolutions, we had to design a \emph{reliable} subjective study, which we ensured by several means. Similar to the work presented in \cite{Hosu2018-expertise-screening}, we invited participants on \texttt{freelancer.com}. The candidates were pre-filtered based on their previous experience, mostly in photography or graphic design, and finally evaluated with regard to their \emph{practical abilities to rate the quality of images}. They had to pass multiple tests in order to qualify for our main study. 

\begin{figure}[b]
    \centering
    \includegraphics[width=\linewidth]{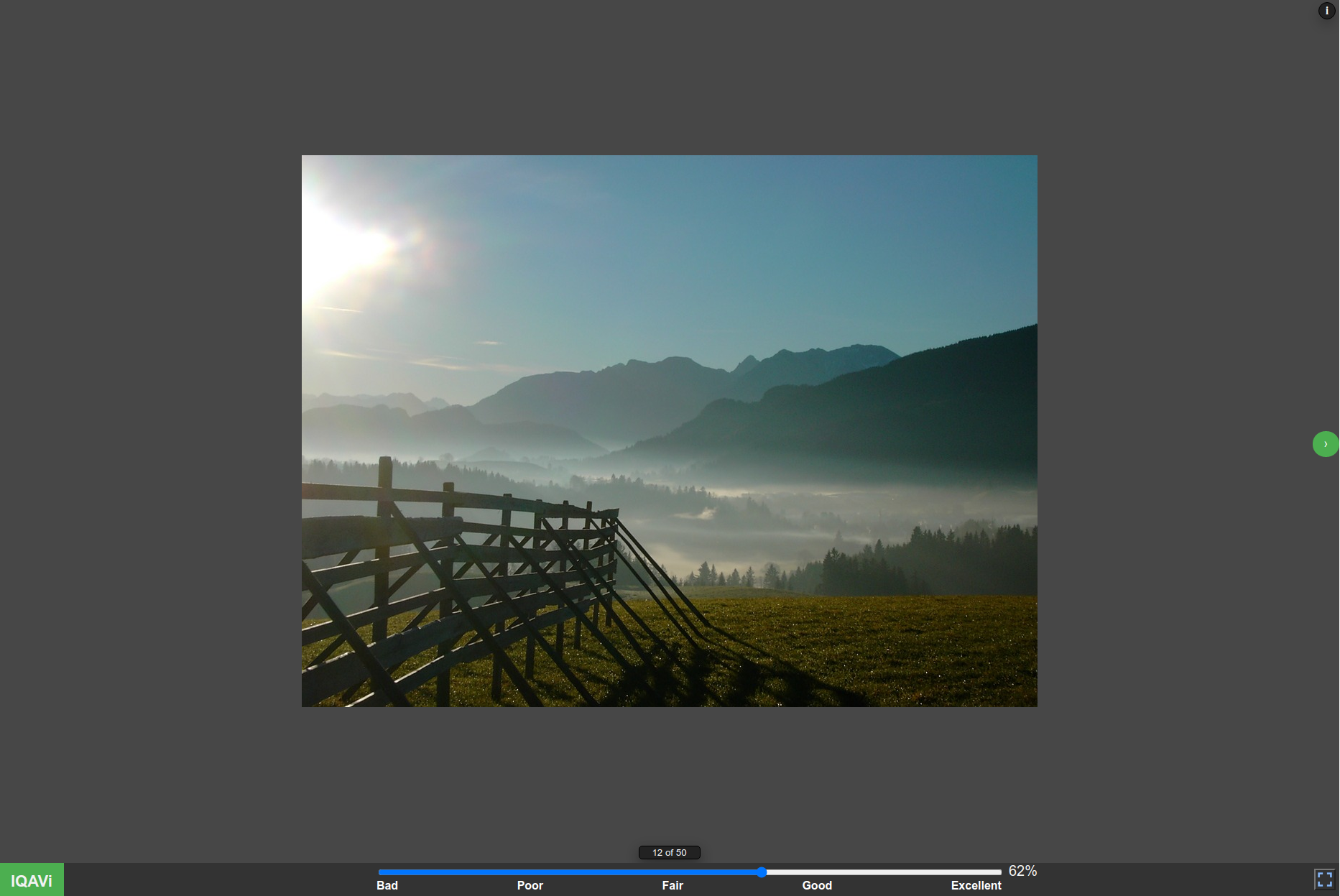}
    \caption{Image quality assessment viewer (IQAVi). }
    \label{fig:iqavi}
\end{figure}

\begin{figure*}[h]
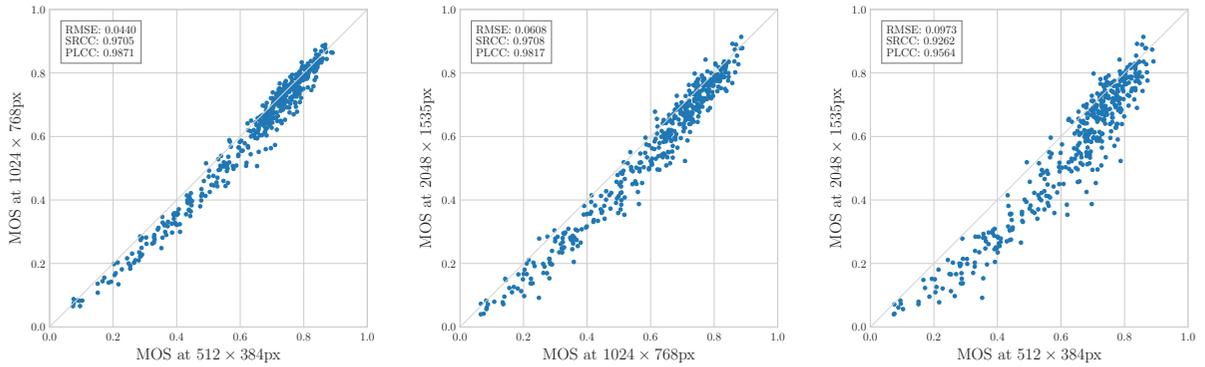

    \begin{center}
        \resizebox{0.32\textwidth}{!}{
        \input{figs/pgf/criq-mos-512-vs-1024.pgf}
        }
        \hfill
        \resizebox{0.32\textwidth}{!}{
        \input{figs/pgf/criq-mos-1024-vs-2048.pgf}
        }
        \hfill
        \resizebox{0.32\textwidth}{!}{
        \input{figs/pgf/criq-mos-512-vs-2048.pgf}
        }
    \end{center}
    \caption{Scatterplots of KonX MOS scores by annotation resolution.
    }
    \label{fig:KonX-res-scatters}
\end{figure*}

\subsubsection{Quality Assessment UI}
We developed a custom web interface that allows to control the image presentation scale and thus enables reproducible studies. It ensures that virtual image pixels are displayed as physical screen pixels in a 1:1 fashion. We account for devices where the \emph{virtual resolution} used in the rendering stage differs from the actual \emph{physical resolution} of the screen.
Ratings were assigned through a slider on a scale from 1 to 100 (\%), which showed labels according to the standard absolute category rating (ACR) scheme. A depiction of our interface is given below in Fig. \ref{fig:iqavi}.

\subsubsection{Participant Filtering}
We conducted a qualifier experiment as a \textit{contest} on \texttt{freelancer.com}. Instructions were given on how to identify distortions, how to judge the overall quality of an image and how to use the rating scale correctly. We carefully explained that judgments should be made independent of the image resolution, as larger presentations are not necessarily better in terms of quality.
We required a screen diagonal size above 14 inches with a resolution of at least $1920\times1080$ pixels and rejected participants with smartphones and small tablets.

While most device checks were fully automated, additional information was gathered through self-reporting from the participants. We stored both the reported and the measured characteristics of all devices that were used in the study.
Participation in a training phase was mandatory for all freelancers. It consisted of 50 images for which we had ground-truth ranges of quality ratings. Upon failing to submit a rating within these bounds we displayed the range of \textit{acceptable} values and users were required to retry until successful.
\begin{figure}[t] % keep this figure within this paragraph or it will move to the next page
    \includegraphics[width=0.925\linewidth]{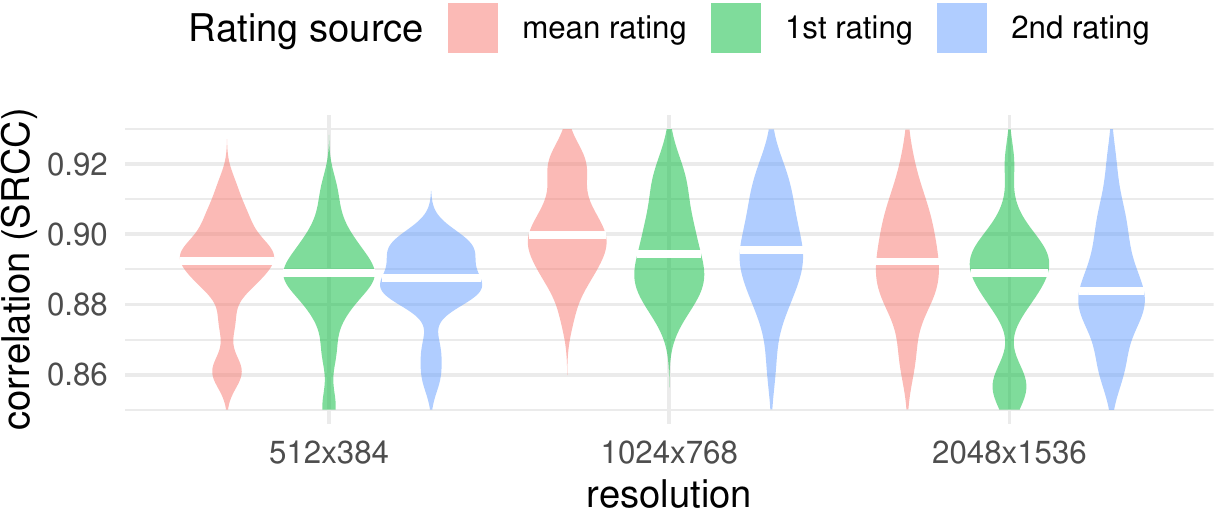}
    \caption{%Correlations between KonX and KonIQ-10k, split by resolution and rating.
    Density of SRCCs of the KonX participant's scores vs the KonIQ-10k MOS. The horizontal white lines indicate their median.}
    \label{fig:correlations-CRIQ-KonIQ-violins}
\end{figure}
%The level of noise exhibited by participants was low to begin with, as shown in Fig. \ref{fig:cor-tries}.
We forced them to keep their browser window maximized during the study. In IQAVi, panning of the currently displayed image allows assessing peripheral content if the image resolution exceeds that of the screen, so those with FHD displays could view the $2048\times 1536$px images in their entirety. We logged the image area in view, as well as the timestamps of annotations and other interactions throughout the experiments for each participant individually.

\subsubsection{Main Annotation Study}
The images in the main study were presented in randomly ordered batches of 50. Each batch contained two repetitions of 25 images of a single resolution. Participants could not check their previous annotations to avoid fraudulent positive effects on their self-consistency. We required them to retry batches on which they failed to meet a SRCC of $0.9$ between their two ratings.

It was rarely necessary to repeat a batch, but when that was the case, almost all batches met the requirements after a single repetition. A participant was asked to repeat a specific batch at most once. The mean of both ratings for an image usually performs better than a single score, as confirmed by computing the correlation to \textit{KonIQ-10k} MOSes (Fig. \ref{fig:correlations-CRIQ-KonIQ-violins}).

\begin{figure}[t]
    \includegraphics[width=\linewidth]{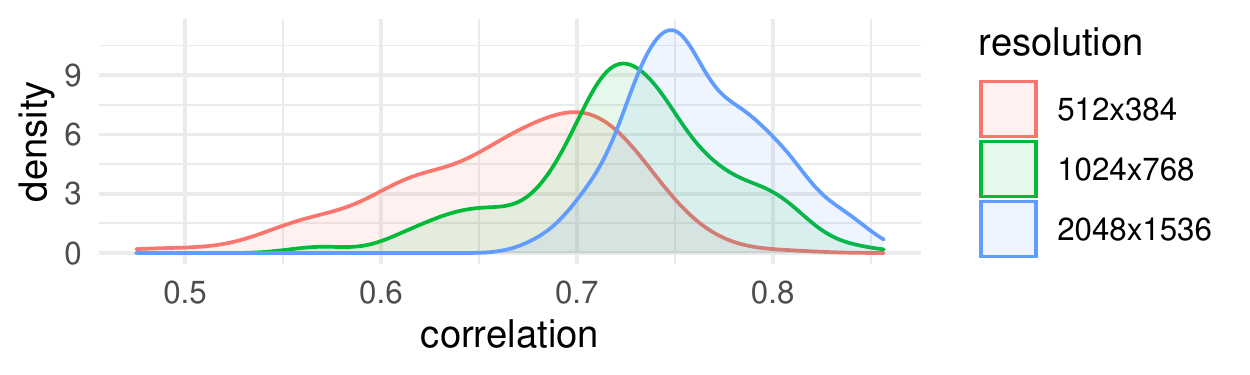}
    \caption{Distribution of SRCCs between all participants in our study, and how they depend on the presented image resolution. Agreements increase with the resolution, indicating that rating the quality of an image is easier at a larger resolution.
    \label{fig:intercorrelations-per-resolution}}
\end{figure}

\subsection{Data Analysis}
\begin{figure}[b]
    \centering
    \includegraphics[width=\linewidth]{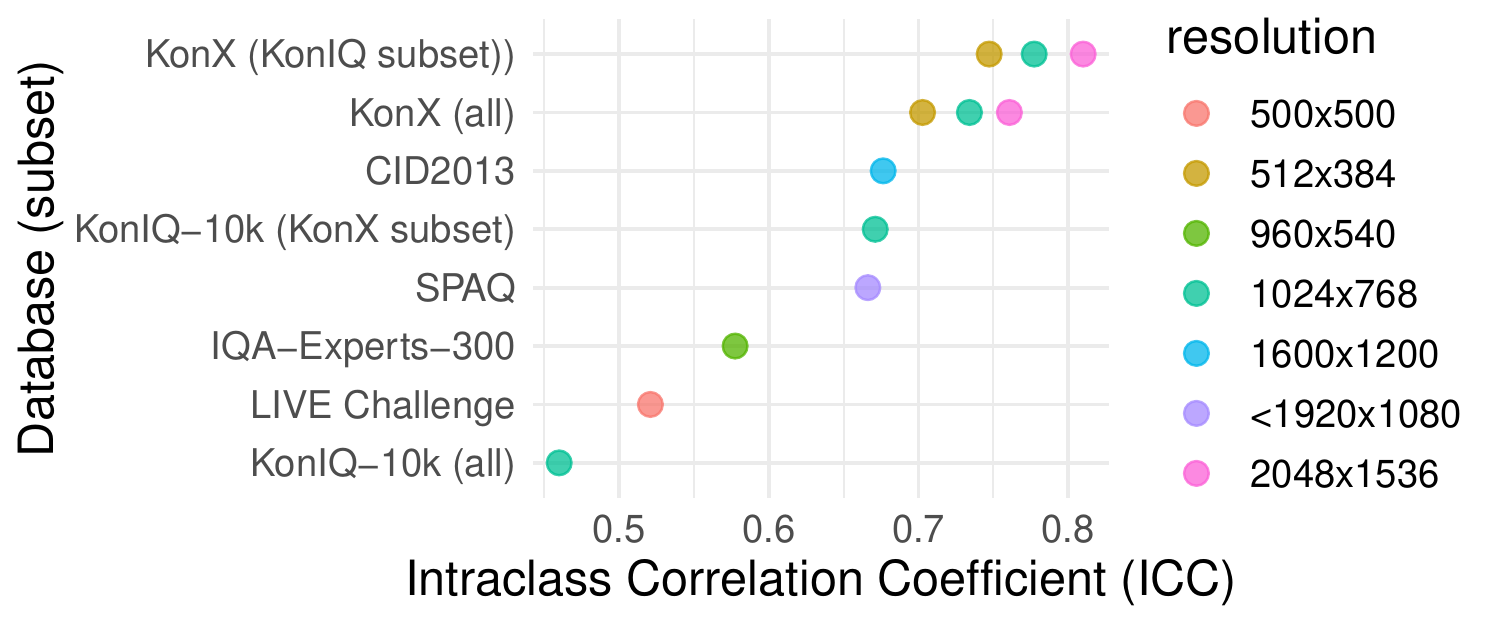}
    \vspace{0.25mm}
    \caption{ICCs \cite{hallgren_computing_2012} for authentically distorted IQA datasets. %, computed using the \texttt{R} package \texttt{ICC::ICCest}.
    For \textit{LIVE Challenge} and \textit{SPAQ} they are approximated based on the MOS and standard-deviations and likely overestimated. The ICC is not always easily comparable across datasets, as it measures the fraction of the total variance accounted for by the per-image (intraclass) variance. Thus, the ICC tends to be larger for databases with a larger spread of the MOS. % The annotation resolution for each dataset was fixed except for SPAQ. % In the latter case, the resolution of the screen was $1920\times1080$. The images were presented proportionally downscaled, the size was not specified by the authors.
    }
    \label{fig:ICCs-databases-more}
\end{figure}

%\subsubsection{Comparison Other Databases}
Reliable, thus reproducible annotations are important for IQA datasets in general, but especially so for KonX due to its primary purpose as a benchmark.
To characterize KonX and to compare it to other datasets, we consider a number of measures. We plot the distribution of inter-user correlations in Fig. \ref{fig:intercorrelations-per-resolution}, measure the intraclass correlation coefficient (ICC) in Fig. \ref{fig:ICCs-databases-more} and investigate the SOS-hypothesis \cite{hossfeld2011sos} in Fig. \ref{fig:sos-plot-authentic}.
The SOS-hypothesis \cite{hossfeld2011sos}  provides an indicator of reliability that accounts for the distribution of MOSes within a dataset. The central point is that the variance of the ratings is constrained by their possible range. If an image MOS is closer to the boundaries of the rating scale, its variance should be smaller than for a MOS at the center of the scale. The $a$ coefficient of a parabola fitted to the variance vs. MOS plot serves as an indicator of reliability.
Larger $a$ means a larger \textit{SOS-normalized variance}, which implies less agreement between ratings. Figure \ref{fig:sos-plot-authentic} shows SOS plots for several databases, including subsets of \textit{KonX} and \textit{KonIQ-10k}.
The ICC(1, 1) coefficient, a one-way random effects single score model \cite{shrout_intraclass_1979,hallgren_computing_2012}, measures the absolute agreement between participants.
This is reasonable, as we have to compare datasets with partial observations.
The ICC is proportional to the variance of the image scores, which is related to the variance of per-image MOSes and roughly inversely proportional to the total variance of all ratings.

It is thus sensible to compare ICCs on the same image subset. For the shared 210 images at $1024\times768$px this indicates improved reliability for KonX over KonIQ-10k, as shown in Fig. \ref{fig:ICCs-databases-more}.
Comparing KonX subsets by resolution suggests that larger images are rated more reliably with better agreement. Furthermore, the inter-user correlations in Fig. \ref{fig:intercorrelations-per-resolution} also indicate that quality assessment might indeed be easier at higher resolutions. This probably is related with the larger difference in quality between the best and the worst images at high resolutions.

\begin{figure}[b]
    \begin{center}
    \resizebox{\linewidth}{!}{
        \input{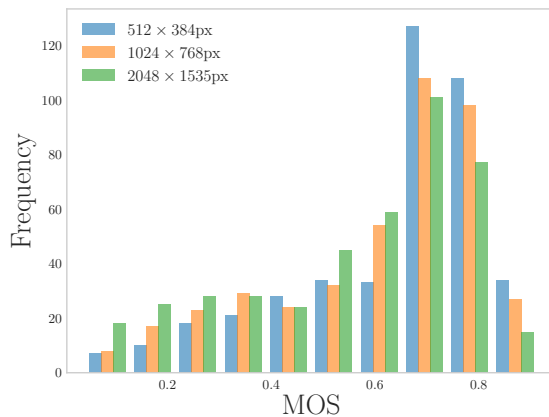}
    }
    \end{center}
    \caption{Histogram of KonX MOS by resolution.}
    \label{fig:criq-mos-histogram}
\end{figure}

\subsubsection{Label Shifts}
We display scatter plots of the MOSes of the same image contents compared by resolution in Fig.~\ref{fig:KonX-res-scatters}. They show curved trends, which match our hypotheses about effects of down-scaling from Section \ref{sec:subjective-factors-in-qoe} quite well.
%High-quality images appear better at a higher resolution, which is indicated by the , represented by markers at the top-right of the plots, appear better at a higher resolution and are thus .
We observe a pronounced preference for the lower resolution in medium quality images, resulting in the shift to the right. There are only few samples at the low-quality end, but the plots indicate that there is a smaller difference in perceptual quality here, i.e. the images look bad regardless of their resolution.

We additionally plot the histograms for the MOS scores per resolution in Fig. \ref{fig:criq-mos-histogram}. 
To formally confirm that there exists a statistically significant difference between the resolution-wise mean opinion scores in KonX we conducted a Wilcoxon signed-rank test for all pairs of resolutions, which is a non-parametric alternative to the popular t-test. The results were significant with $p < 0.005$ for all pairs.

\subsubsection{Summary}

We conclude from this analysis that KonX is reliably annotated, especially in contrast to previous works. This is likely due to multiple factors, including the following design choices we made:

\begin{enumerate}[label=\roman*), topsep=\baselineskip]
\item Usage of a fine-grained annotation scale instead of the traditional five-point ACR.
\item Consistency checks of the participants, as all items were repeated twice in the study.
\item Noise-reduction by averaging the repetitions for each participant individually.
\item A high(er) level of control, especially by rendering image pixels 1:1 to screen pixels.
\end{enumerate}

\section{Cross-Resolution Prediction} % and Model Architecture}
\label{sec:approaches}

Our model architecture is inspired by several observations from the literature regarding the properties of features from different CNN layers, their scale dependence, and their effect on transfer learning. Scale-dependence is obvious for individual filters, meaning that they can only detect patterns of a fixed size. This is less evident for groups of filters or the usual cascades of convolutions used in deep CNNs. %Nonetheless, scale-specific representations are present deep in neural architectures.
ImageNet models for example achieve a certain degree of scale-invariance of object classes only close to the last layers \cite{graziani2021ScaleInvarianceState}. We considered multiple aspects:\\

%\noindent
\textbf{Train-Test Scale Discrepancy:} Object classification models that were trained closer to the test resolutions perform better after fine-tuning, which we expect to hold for IQA as well \cite{touvron2019FixingTraintestResolution}. 

%\noindent
\textbf{Scale-Agnostic Features:} Following the observations of Graziani et al. \cite{graziani2021ScaleInvarianceState} on scale-invariance, the prevalent use of late-stage features could be suboptimal for quality assessment.

%\noindent
\textbf{Multi-Level Binding:} The connection between the backbone and head network is traditionally based on the outputs of a single late-stage layer. Cross-task learning might be limited by this, as the success of multi-level features in well-performing architectures \cite{hosu2019effective,li_norm-in-norm_2020} suggests.

%\noindent
\textbf{Resolution Overfitting:} Modern DNN architectures for NR-IQA accept one input size at a time. We found in our limited experiments that training such models on multiple resolutions did not improve their cross-resolution performance, on the contrary, it often decreased it. Learning scale-specific features on only one common network architecture seems to be a limitation of this approach, at least in practice with limited time and training data.

\subsection{NR-IQA Model Architecture}
To get around these difficulties with our architecture we made the following design choices:

\begin{itemize}[topsep=\baselineskip]
    \item An EfficientNet-B7 \cite{tan2019efficientnet} pre-trained at $600\times600$px serves as a backbone, which is close to our targeted resolutions and has been shown to be tweakable regarding input scales \cite{touvron2020fixing}.
    \item The Inception-MLSP approach from \cite{hosu2019effective} gets adapted to EfficientNet by substituting Inception-module output activations with an inner layer of the EfficientNet-modules.
    \item We train a two-column network, similar to those used for scale-invariant detection \cite{Kang2018crowd,zhang2016SingleImageCrowdCounting,walach2016LearningCountCNN,onororubio2016Perspective}, at different input resolutions. This enables the deep integration of column-wise MLSP-type features, synergizing with the proposed shallow-binding fix.
\end{itemize}

\begin{figure*}[t]
  \begin{center}
  \centering
  \hfill
  \includegraphics[width=0.55\linewidth]{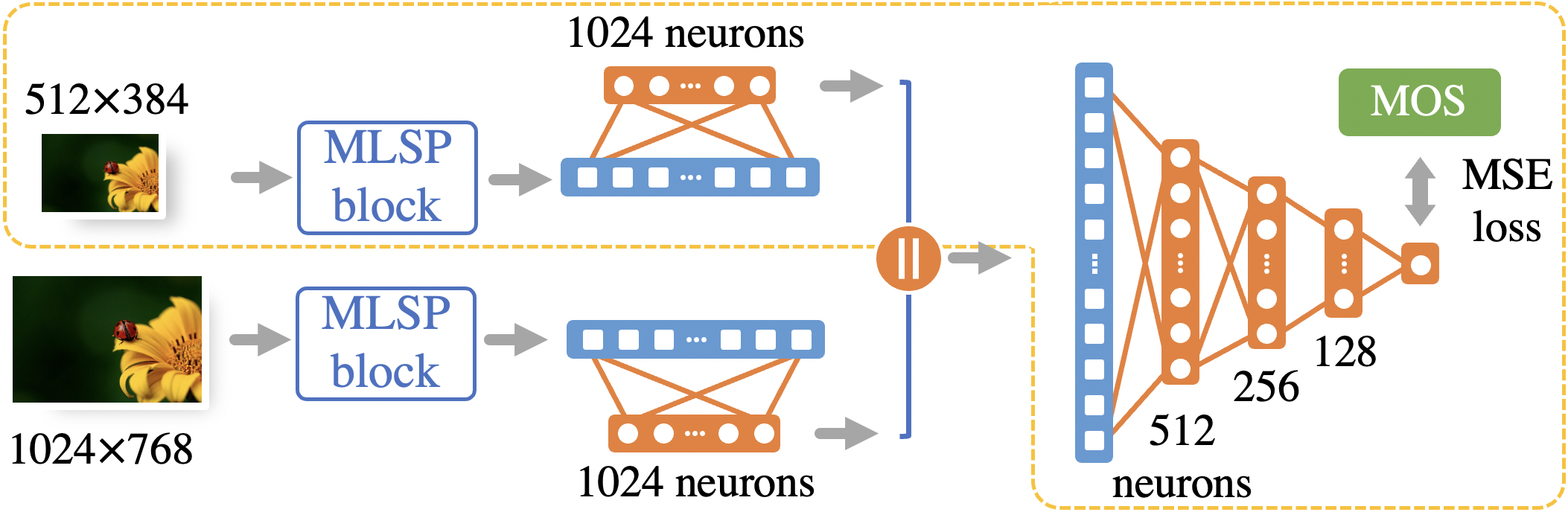}
  \hfill
  \includegraphics[width=0.35\linewidth]{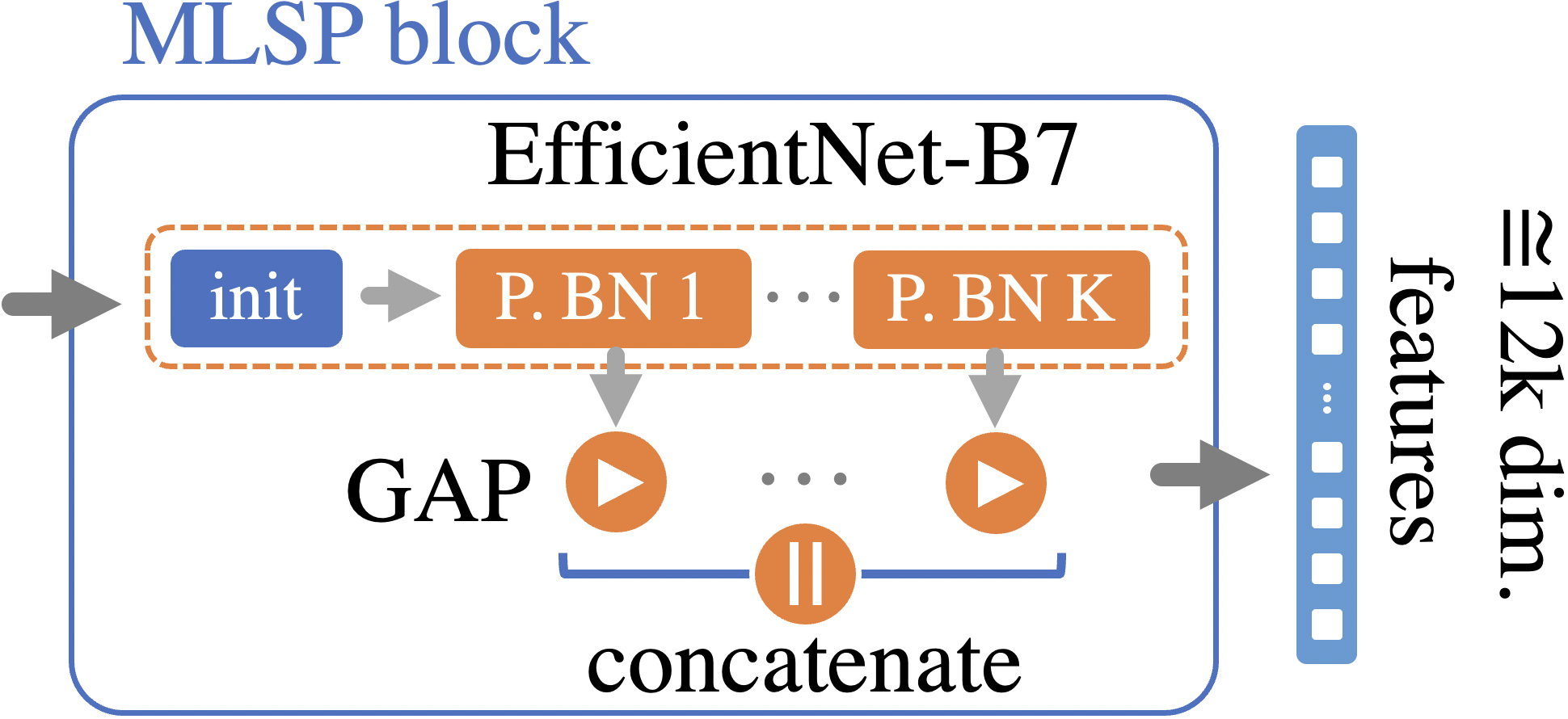}
  \hfill
  \end{center}
  \vspace{3mm} % graphics themselves have no surrounding white-space, hope it's okay to add some here
  \caption{The proposed \texttt{Effnet-2C-MLSP} two-column NR-IQA architecture. The yellow-dotted section on the left figure describes the single-column (\texttt{1C}) variants, \texttt{P.BN K} refers to the  \texttt{project\_bn} layers.
  }
  \label{fig:efficientnet-mlsp-architecture}
\end{figure*}

The proposed \texttt{Effnet-2C-MLSP} is depicted in Fig. \ref{fig:efficientnet-mlsp-architecture}. It consists of two columns (2C) of MLSP \cite{hosu2019effective} blocks based on independent-weights EfficientNet-B7 backbones. These were pre-trained on ImageNet-1000 at $600\times600$px as a  middle ground for the fine-tuning at $512\times384$px and $1024\times768$px.
%We aim to generalize to resolutions as high as $2048\times1536$px.

Both columns feed into a cascaded multi-layer-perceptron (MLP) head. Features are sampled by global average pooling (GAP) the activations of the \texttt{project\_bn} layers; this is different from Inception-MLSP features \cite{hosu2019effective,lin2019kadid} which stem from \textit{mixed} layers. Their analog in ResNet-architectures would be the \texttt{add} layers at the end of each module, which are redundant due to the residual connections. Since the immediately preceding layers use dropout normalization, we extract the outputs from two layers before. In our preliminary experiments neither the \texttt{add} nor the \texttt{dropout} activations performed better.

The \texttt{project\_bn} features contain about $12000$ scalar values, which we downsize to $1024$ through separate dense layers for each column before passing them to the MLP head; the downsizing significantly reduces the number of parameters needed. This hierarchical combination allows for a greater level of per-scale differentiation of the column features through backpropagation compared to simply adding the features together. The models are trained to predict a single mean opinion score (MOS) directly, steered by the MSE loss. 

\subsection{Training Data}
\textit{KonX} is now available as a test set, but there is no cross-resolution equivalent that is sufficiently large for training.
%To the best of our knowledge, KonIQ-10k \cite{hosu2020koniq} is the authentically distorted dataset that was annotated at the highest resolution of $1024\times768$px. Other studies were performed at lower resolutions, mostly below $800\times800$px \cite{Yumin2020Perceptual-SPAQ,ghadiyaram2015live,ying2020patches}.
Existing datasets \cite{hosu2020koniq,ghadiyaram2015live,Yumin2020Perceptual-SPAQ,ying2020patches}, for which each image was presented for rating at a single resolution \footnote{Paq-2-Piq \cite{ying2020patches} patches have to be considered as entirely different images because the placement of the patch sampling affects their perceptual quality.} limit training to this respective annotation resolution. We can mitigate this shortcoming by exploiting a data overlap.

\begin{figure*}[t]
    \centering
    \includegraphics[width=0.29\textwidth]{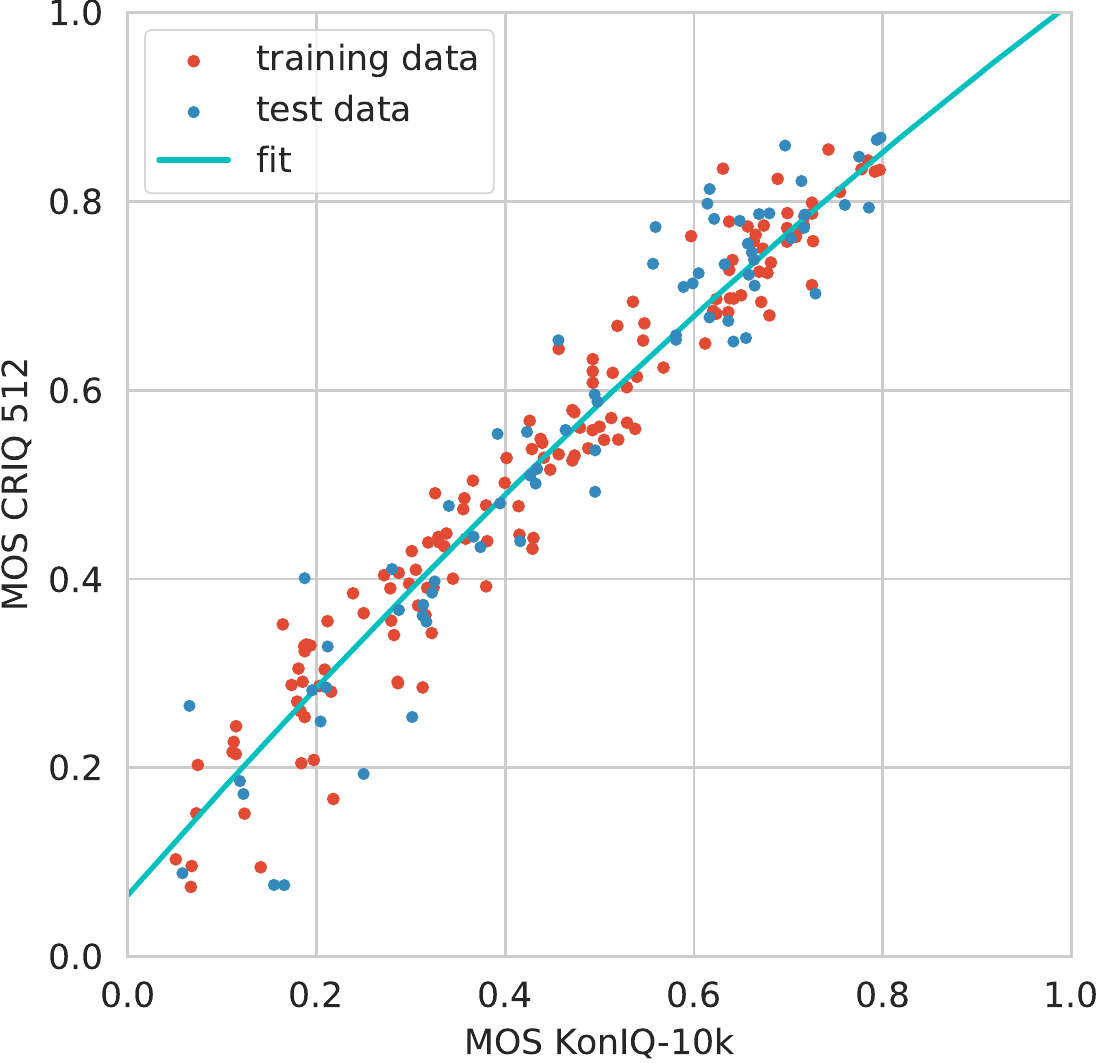} \hfill
    \includegraphics[width=0.29\textwidth]{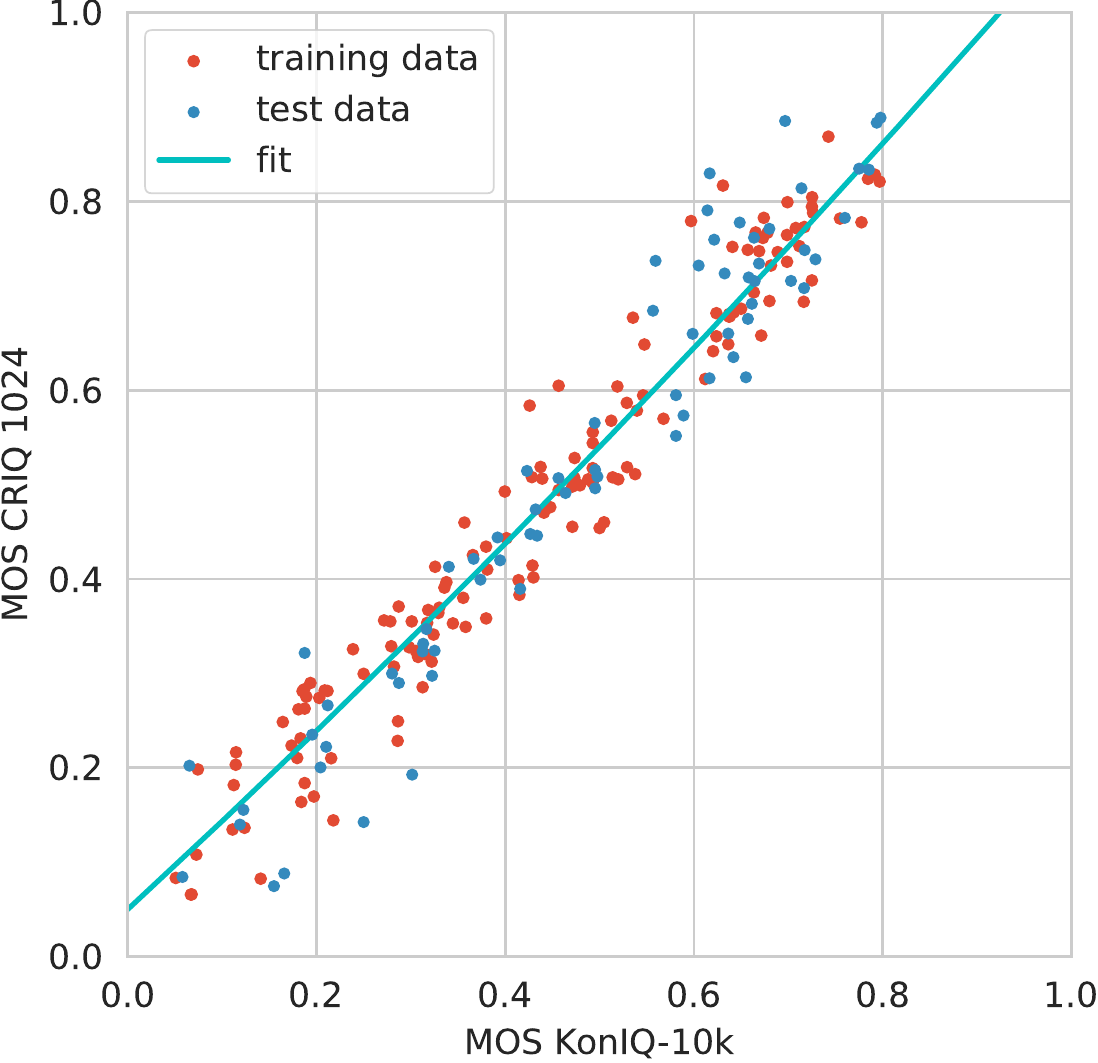} \hfill
    \includegraphics[width=0.29\textwidth]{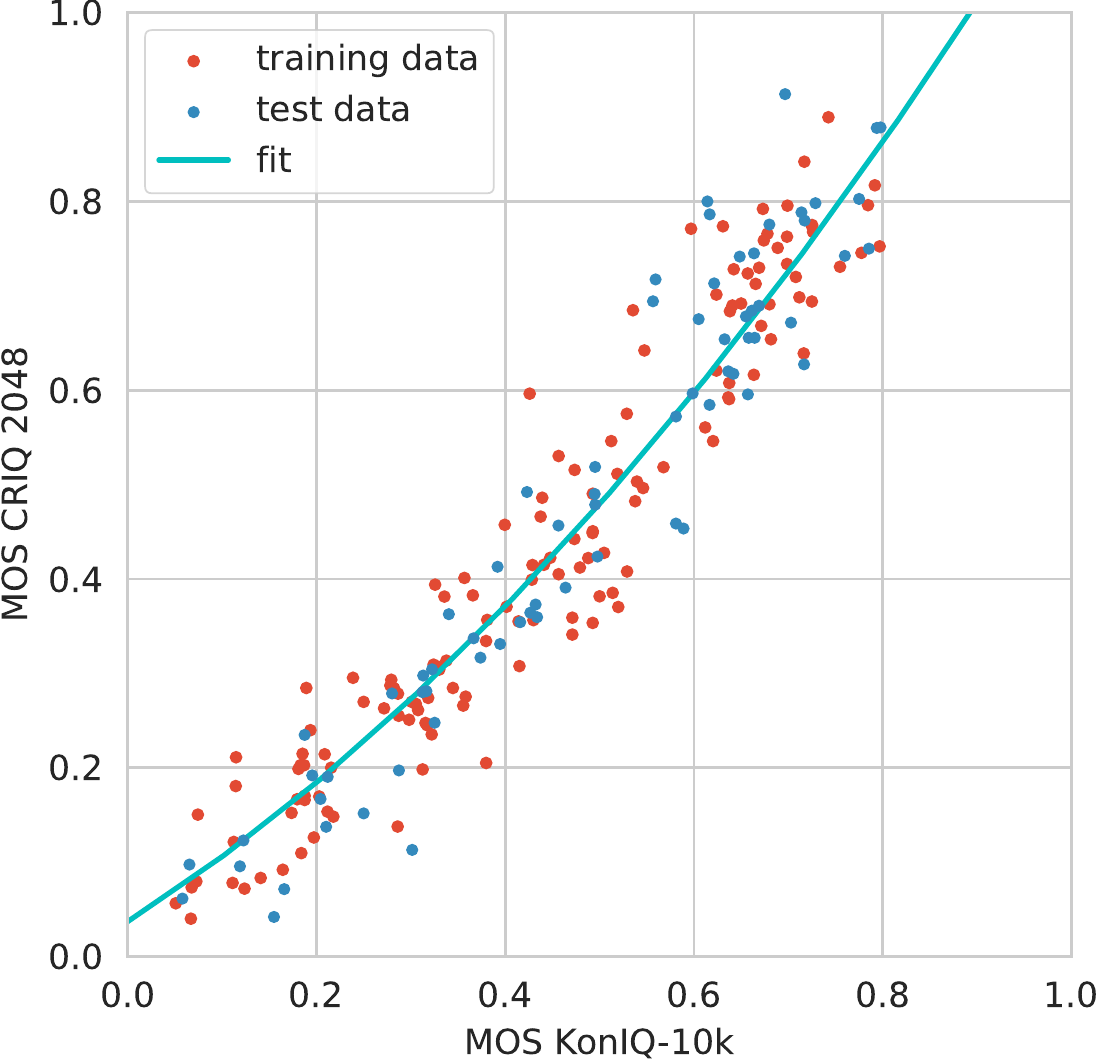}
    \vspace{1mm}
    \caption{Quadratic mapping from KonIQ-10k MOS to KonX at all three resolutions to align the scores for training at different resolutions on KonIQ-10k and evaluation on KonX. The blue markers were kept as a test-set to determine the quality of the fit. At $1024\times 768$px the scores are essentially just shifted.
    \label{fig:fit}}
\end{figure*}

Fitting quadratic functions that map MOS scores from \textit{KonIQ-10k} to each of the resolutions in \textit{KonX} allows to align the scores between datasets and resolutions.
We propose this as a better approximation of the underlying ground-truth labels than using the KonIQ-10k \footnote{KonIQ-10k was annotated at $1024\times 768$px.} scores for different resolutions directly. This adaptation reduces the MAE by 12.8\% and the MSE by 20.3\% over all three resolutions, as determined on a test-set of 70 images that were not utilized in the curve fitting,
as shown in Fig. \ref{fig:fit}.

We excluded the 210 images sampled for \textit{KonX} from \textit{KonIQ-10k} and created a $5$-fold train/test split with the property that one of the test sets is a subset of the original \textit{KonIQ-10k} test set. Each model under consideration is trained and evaluated on all folds. We report performance indicators for each \textit{KonX} subset in Table \ref{tab:results_KonX} and show cross-test results on other datasets in Table \ref{tab:crosstests}.

\subsubsection{Training Strategy}
Training of \texttt{Effnet-2C-MLSP} was conducted in two stages. First, we kept the weights of the MLSP blocks fixed and trained just the head. This already achieves close to optimal performance and converges fast. In the second stage, we fine-tuned both columns jointly, but did not update the batch normalization layers. Each stage is run for at most 40 epochs, with early stopping in 10 epochs if the validation loss does not improve.

The learning rates for the two stages were $10^{-5}$ and $10^{-4}$, respectively.
Incrementally fine-tuning one column at a time resulted in inferior results. The only augmentation we used was horizontal flipping of images, doing this independently per column improved performance marginally. We feed the entire image at a time. %, rescaled as needed.
In our experiments, cropping the images  did not provide a performance improvement.

Initial experiments with the Adam and SGD optimizers lead to unsatisfactory performance. The large resolutions and small batch sizes caused divergence, and the training loss increased rapidly after the first few epochs of the second stage. In order to reduce the effect of large gradients, we used gradient clipping (\texttt{clipnorm}=$1.0$), which worked well. We ultimately switched to the NAdam \cite{dozat2016incorporating} optimizer with Nesterov momentum.

\subsubsection{Model Performance Evaluation}

\begin{table*}[b]
\centering
\resizebox{0.65\linewidth}{!}{
\begin{tabular}{@{}lcccccc@{}}
\toprule
Models         & \multicolumn{2}{c}{KonIQ-10k} & \multicolumn{2}{c}{Live Challenge} & \multicolumn{2}{c}{SPAQ} \\ \midrule
               & SRCC         & PLCC          & SRCC            & PLCC            & SRCC             & PLCC             \\
LinearityIQA   & 0.9299          & 0.9415        & 0.8114           & 0.8404          & 0.8442            & 0.8422           \\
Effnet-NIMA    & 0.7635          & 0.7788        & 0.6886           & 0.7269          & 0.7896            & 0.7936           \\
IRN-1C-MLSP    & 0.8601          & 0.8932        & 0.8005           & 0.8310          & 0.8523            & 0.8553           \\
Effnet-2C-MLSP & \textbf{0.9490} & \textbf{0.9596} & \textbf{0.8327} & \textbf{0.8595} & \textbf{0.8641} & \textbf{0.8641}           \\ \bottomrule
\end{tabular}
}
\caption{Cross database tests: training was conducted on KonIQ-10k, testing on the respective datasets.% Our proposed \texttt{Effnet-2C-MLSP} performs best across all databases and metrics.
}
\label{tab:crosstests}
\end{table*}

Our \texttt{Effnet-2C-MLSP} was evaluated by feeding each column a different version of the same image: For the low-resolution column, images were always resized to $512\times384$px. The other column received the original image size. When testing on e.g. $2048\times1536$px KonX images, a downscaled $512\times384$px version was presented to the low-resolution column, and the $2048\times1536$px original to the other one.
We cross-validated on 5-folds. The test sets are non-overlapping. The training database used was the remapped KonIQ-10k, after removing the 210 images that are shared with KonX. Thus, each set (training, validation, and test) is slightly smaller than the official splits published for KonIQ-10k.

\begin{table*}[t]
\resizebox{\textwidth}{!}{

\begin{tabular}{@{}lccccccccccccc@{}}
\toprule
\textbf{Model}                  & \textbf{\begin{tabular}[c]{@{}c@{}}Training\\ Resolution\end{tabular}} & \multicolumn{6}{c}{\textbf{SRCC}}                                                                                & \multicolumn{6}{c}{\textbf{PLCC}}                                                                                 \\ \midrule
                                &                                                                        & \multicolumn{2}{c}{$512\times384$px} & \multicolumn{2}{c}{$1024\times768$} & \multicolumn{2}{c}{$2048\times1536$} & \multicolumn{2}{c}{$512\times384$px} & \multicolumn{2}{c}{$1024\times768$} & \multicolumn{2}{c}{$2048\times1536$} \\
                                &                                                                        & KonIQ             & Pixabay          & KonIQ            & Pixabay          & Koniq             & Pixabay          & Koniq             & Pixabay          & Koniq            & Pixabay          & Koniq             & Pixabay          \\ \cmidrule(l){3-14}
\multirow{2}{*}{KonCept}        & 512                                                                    & 0.8807            & 0.3047           & 0.8264           & 0.2703           & 0.6821            & 0.3112           & 0.8535            & 0.3049           & 0.7522           & 0.2670           & 0.6016            & 0.2690           \\
                                & 1024                                                                   & 0.8251            & 0.2658           & 0.8888           & 0.4175           & 0.8165            & 0.4518           & 0.6968            & 0.2658           & 0.8845           & 0.4201           & 0.8420            & 0.4926           \\ \hline
\multirow{2}{*}{Effnet-NIMA}           & 512                                                                    & 0.8506            & 0.3101           & 0.7648           & 0.3739           & 0.5505            & 0.4010           & 0.8357            & 0.3682           & 0.7664           & 0.4118           & 0.5928            & 0.3972           \\
                                & 1024                                                                   & 0.8568            & 0.2506           & 0.8840           & 0.3184           & 0.8185            & 0.3925           & 0.8449            & 0.3105           & 0.8849           & 0.3895           & 0.8423            & 0.4503           \\ \hline
\multirow{2}{*}{LinearityIQA}   & 512                                                                    & \textbf{0.9436}   & 0.3818           & 0.9111           & 0.3994           & 0.7611            & 0.4485           & \textbf{0.9416}   & 0.4681           & 0.9068           & 0.4670           & 0.7933            & 0.4859           \\
                                & 1024                                                                   & 0.9141            & 0.3849           & \textbf{0.9452}  & 0.4519           & 0.9023            & 0.4935           & 0.9087            & 0.4311           & 0.9435           & 0.4813           & 0.9115            & 0.5291           \\ \hline
\multirow{2}{*}{IRN-1C-MLSP}    & 512                                                                    & 0.9279            & 0.3197           & 0.9093           & 0.3490           & 0.8072            & 0.4501           & 0.9274            & 0.4155           & 0.9046           & 0.4355           & 0.8326            & 0.4967           \\
                                & 1024                                                                   & 0.8949            & 0.3117           & 0.9320           & 0.4190           & 0.9076            & 0.5037           & 0.8992            & 0.4003           & 0.9313           & 0.4876           & 0.9160            & \textbf{0.5596}  \\ \hline
\multirow{3}{*}{Effnet-2C-MLSP} & 512                                                                    & 0.9273            & 0.3955           & 0.9056           & 0.4457           & 0.7900            & 0.5149           & 0.9248            & 0.4689           & 0.9035           & 0.5063           & 0.8252            & 0.5391           \\
                                & 1024                                                                   & 0.8918            & 0.3762           & 0.9358           & 0.4844           & 0.9105            & \textbf{0.5415}  & 0.8957            & 0.4443           & 0.9361           & \textbf{0.5422}  & 0.9228            & 0.5857           \\
                                & both                                                                   & 0.9234            & \textbf{0.4058}  & 0.9426           & \textbf{0.4715}  & \textbf{0.9276}   & 0.5132           & 0.9251            & \textbf{0.4783}  & \textbf{0.9437}  & 0.5220           & \textbf{0.9325}   & \textbf{0.5596}  \\ \bottomrule
\end{tabular}
}%resizebox
\caption{Correlations on KonX subsets when training and testing at different resolutions. SRCC and PLCC is the Spearman's Rank and Pearson linear correlation coefficient.} 
\label{tab:results_KonX}
\end{table*}

%\subsection{Model Performance Comparison}
We compare to previous works on \textit{KonX} and the \textit{KonIQ-10k} \cite{koniq10k} test set as well as in cross-tests on LIVE-ITW \cite{ghadiyaram2015live} and SPAQ \cite{Yumin2020Perceptual-SPAQ}. Table \ref{tab:results_KonX} shows correlations per subset, split by training and test resolution as well as data source. 
We trained and tested KonCept-512 \cite{koniq10k}, LinearityIQA \cite{li_norm-in-norm_2020} and an
\textit{EfficientNet}-based derivative (ours) of NIMA \cite{talebi2018nima} for an up to date comparison.

\begin{figure*}[b]
    \centering
    \includegraphics[width=0.6\linewidth]{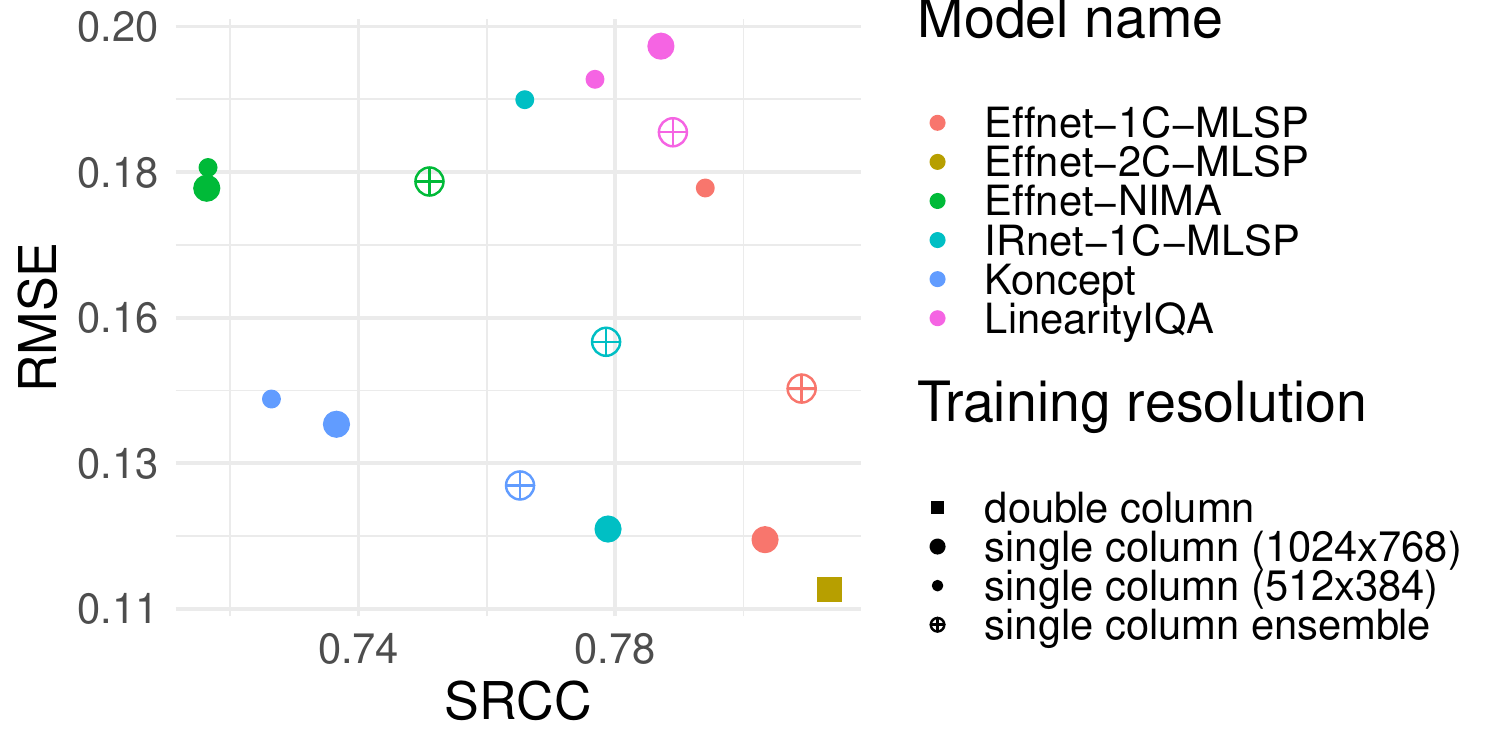}
    \caption{RMSE vs. rank correlations (SRCC) were calculated jointly over the entirety of KonX on all resolutions. We report averages over all five splits. Through the RMSE, a key indicator of cross-resolution performance, this plot reveals biased but highly correlated predictions. We also report single resolution/column performance and that of ensembles made of two single-column predictors where the individual model's outputs are averaged. Our \texttt{Effnet-2C-MLSP} has the highest SRCC and lowest RMSE.}
    \label{fig:correlation_vs_error}
\end{figure*}

An ablation study on the backbone network selection is included in the table. The \textit{EfficientNet-B7} was replaced in \texttt{IRN-2C-MLSP} with an \textit{InceptionResNetV2}, which, as previously stated, was successfully used in many IQA related experiments. As suggested by Fig. \ref{fig:gradcam}, this architecture suffers from cross-resolution discrepancies and is indeed outperformed by the \textit{EfficientNet}-based architecture. 
An overview of the SRCC and MSE performances is given in Fig. \ref{fig:correlation_vs_error}, which shows that \texttt{Effnet-2C-MLSP} is highly performant, with respect to its accuracy and correlations with the ground-truth. \texttt{Effnet-2C-MLSP} also performs best when evaluated against the KonIQ-10k test set and across test sets on Live ITW and SPAQ (at $1920\times 1080$px) as shown in Table \ref{tab:crosstests}. Absolute error metrics (MSE) are crucial on KonX. The concentration of images at the top of the quality scale results in lower correlations on the Pixabay subset, making it more difficult to distinguish model performances. Nonetheless, our proposed model excels on both metrics.

\section{Conclusions}
\label{sec:conclusions}
This paper introduced the cross-resolution NR-IQA problem, which is a step toward assessing modern high-resolution images with computer vision models.
We made significant progress in predicting the quality of authentically distorted images of various sizes. 
For that purpose, we introduced \textit{KonX}, a benchmark dataset crafted specifically for cross-resolution IQA.

It includes 420 images from two source domains and is reliably annotated at three presentation resolutions through a subjective study. For the first time, the database allows for the study of the effects of cross-resolution independent of cross-content, while also allowing for cross-domain experiments by splitting on the data source. 
We additionally established a solid foundation for cross-resolution prediction with our \texttt{Effnet-2C-MLSP} model, which achieves state-of-the-art performance also across databases.

As auxiliary results, we tapped into the importance of the pre-training resolution relative to the post-fine-tuning performance regarding scale-overfitting, the usage of multi-level features with varying levels of scale-variance and the application of column-wise multi-scale training in IQA.
Considering these aspects surely helped, but they are far from being completely understood.
Our work thus opens up new avenues for research in this field, such as developing computationally less intensive architectures and adapting  advances in IQA to video quality assessment.

\subsubsection*{Acknowledgements}
Funded by the Deutsche Forschungsgemeinschaft (DFG, German Research Foundation) – Project-ID 251654672 – TRR 161 % As per Leonels mail from 4/2/21

\bibliography{literature}% common bib file
%% if required, the content of .bbl file can be included here once bbl is generated
%%\input sn-article.bbl

%% Default %%
%%\input sn-sample-bib.tex%

\end{document}